\documentclass[sigconf]{acmart}

\AtBeginDocument{%
  }

\usepackage{microtype}

\usepackage{algorithm}
\usepackage{algpseudocode}
\usepackage{array}
\usepackage{enumitem} 
\usepackage{booktabs}
\usepackage{multirow}
\usepackage{amsmath}
\usepackage{siunitx}
\usepackage{makecell}
\usepackage{xcolor}
\usepackage{colortbl}
\usepackage{graphicx} 
\usepackage{soul} 
\usepackage{wrapfig}
\usepackage{subcaption}
\usepackage{appendix}
\usepackage{cleveref}
\usepackage{balance}

\newcommand{\graycell}[1]{\cellcolor{gray!20}#1}
\newcommand{\hlc}[2]{{%
    \setlength{\fboxsep}{1.0pt}
    \colorbox{#1}{$\mathstrut #2$}%
}}
\definecolor{bgyellow}{RGB}{255, 245, 230}
\definecolor{bgblue}{RGB}{228, 248, 253}
\definecolor{bggreen}{RGB}{245, 251, 232}
\definecolor{bgred}{RGB}{192, 0, 0}
\definecolor{sciblue}{RGB}{200,225,245}
\definecolor{linkpink}{RGB}{210, 70, 130}
\sethlcolor{sciblue}

\algnewcommand{\Input}[1]{\State \textbf{Input:} #1}
\algnewcommand{\Output}[1]{\State \textbf{Output:} #1}
\algnewcommand{\Hyperparams}[1]{\State \textbf{Hyperparameters:} #1}
\algtext*{EndIf}
\algtext*{EndFor}
\algtext*{EndWhile}
\algtext*{EndProcedure}
\AtBeginDocument{%
  \setlength{\abovedisplayskip}{3pt plus 1pt minus 1pt}
  \setlength{\belowdisplayskip}{3pt plus 1pt minus 1pt}
  \setlength{\abovedisplayshortskip}{1pt plus 1pt minus 1pt}
  \setlength{\belowdisplayshortskip}{2pt plus 1pt minus 1pt}
}
\AtBeginDocument{%
  \setlength{\textfloatsep}{7pt plus 1pt minus 2pt}
  \setlength{\floatsep}{6pt plus 1pt minus 1pt}
  \setlength{\intextsep}{6pt plus 1pt minus 2pt}
  \setlength{\dbltextfloatsep}{7pt plus 1pt minus 2pt}
  \setlength{\dblfloatsep}{6pt plus 1pt minus 1pt}
  \setlength{\abovecaptionskip}{3pt}
  \setlength{\belowcaptionskip}{2pt}
}
\copyrightyear{2026}
\acmYear{2026}
\setcopyright{cc}
\setcctype{by}
\acmConference[MM '26] {Proceedings of the 34th ACM International Conference on Multimedia}{November 10--14, 2026}{Rio de Janeiro, Brazil.}
\acmBooktitle{Proceedings of the 34th ACM International Conference on Multimedia (MM '26), November 10--14, 2026, Rio de Janeiro, Brazil}
\acmISBN{979-8-4007-2213-4/2026/11}
\acmDOI{10.1145/3767308.3836455}

\begin{document}

\title{VARPose: Flexible 2D Pose Densification via Visual Autoregressive Modeling for Enhanced 3D Lifting}

\author{Kaiyuan Pu}
\authornote{Both authors contributed equally to this research.}
\affiliation{%
  \institution{School of Artificial Intelligence, \\Sun Yat-sen University}
  \city{Zhuhai}
  \country{China}}
\email{puky@mail2.sysu.edu.cn}
\author{Tiantian Yang}
\authornotemark[1]
\affiliation{%
  \institution{School of Artificial Intelligence, \\Sun Yat-sen University}
  \city{Zhuhai}
  \country{China}}
\email{yangtt69@mail2.sysu.edu.cn}
\author{Dan Zeng}
\authornote{Corresponding author: zengd8@mail.sysu.edu.cn.}
\affiliation{%
  \institution{School of Artificial Intelligence, \\Sun Yat-sen University}
  \city{Zhuhai}
  \country{China}}
\affiliation{%
  \institution{The Technology Innovation Center for Collaborative Applications of Natural Resources Data in GBA, MNR}
  \city{Guangzhou}
  \country{China}}
\email{zengd8@mail.sysu.edu.cn}

\begin{abstract}
  Visual AutoRegressive Modeling (VAR) has excelled in natural image generation via next-scale prediction, but its use on topology-structured data like human skeletons is still unexplored. VARPose is proposed to adaptively densify 2D sparse poses, thereby enriching the anatomical information available for 3D lifting models. Our core contributions are twofold. First, we introduce a Granularity-agnostic Pose Tokenizer (GPT), which employs a single hybrid codebook and a residual quantization strategy to encode poses of varying densities into a unified, multi-scale discrete representation. Our results demonstrate the strong generalizability of this representation. By decoupling the representation from the projection, we can successfully decode novel pose granularities using a frozen codebook with a retrained decoder. Second, we propose UniSkelar, a unified autoregressive model that treats ``joint density'' as ``scale''. UniSkelar learns to predict the token sequence for the next density level in a coarse-to-fine manner, conditioned on the sparsest pose. VARPose not only outperforms state-of-the-art methods and generalizes to unseen granularities, but also confers tangible performance gains on downstream tasks, such as 3D Pose Estimation and Human Mesh Recovery, through 2D pose densification. Our code and model are available at \textcolor{purple}{\url{https://github.com/BRL-SYSU/VARPose.git}}.
\end{abstract}



\begin{teaserfigure}
  \includegraphics[width=\textwidth]{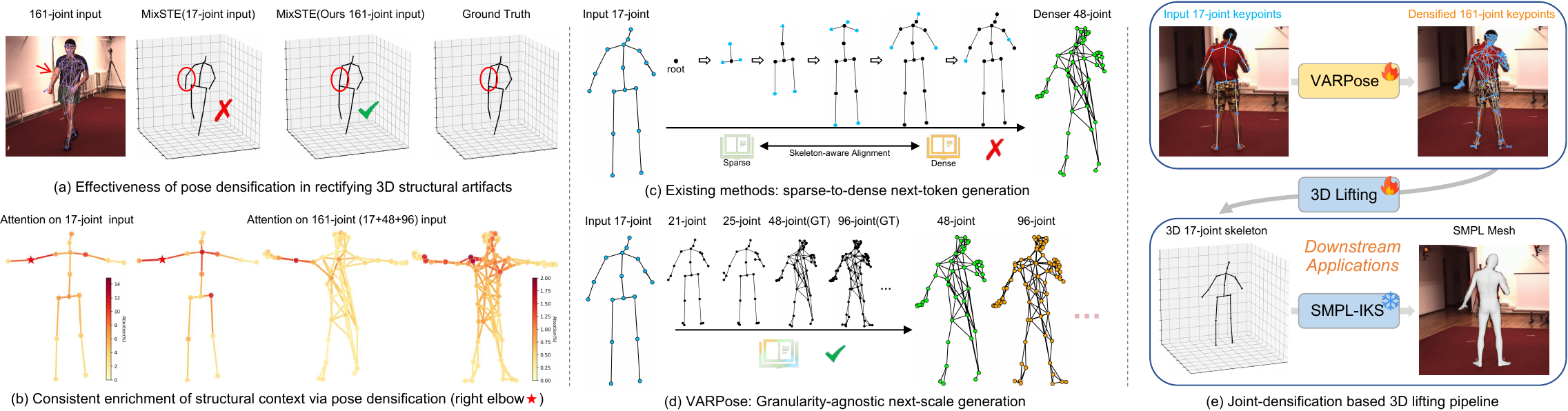}
  \caption{Comparison of pose generation paradigms and our downstream lifting pipeline. (a--b) 3D results and attention heatmaps demonstrate that pose densification enriches structural context, effectively correcting errors in 3D lifting. (c) Existing methods rely on local next-token prediction with separate codebooks for fixed granularities. (d) VARPose leverages a unified codebook and next-scale prediction to introduce a novel paradigm for synthesizing poses with flexible granularities. (e) Our pipeline translates sparse 2D inputs into densified poses, accurate 3D skeletons, and SMPL meshes.}
  \Description{Comparison of pose generation paradigms and our downstream lifting pipeline. (a--b) 3D results and attention heatmaps demonstrate that pose densification enriches structural context, effectively correcting errors in 3D lifting. (c) Existing methods rely on local next-token prediction with separate codebooks for fixed granularities. (d) VARPose leverages a unified codebook and next-scale prediction to introduce a novel paradigm for synthesizing poses with flexible granularities. (e) Our pipeline translates sparse 2D inputs into densified poses, accurate 3D skeletons, and SMPL meshes.}
  \label{fig:varpose_frameworks}
\end{teaserfigure}

\maketitle

\section{Introduction}
3D human pose estimation (HPE) seeks to accurately predict the 3D coordinates of human skeletal joints from monocular images or videos. This capability is crucial as it provides a deep understanding of complex human behavior and has broad applications such as human-computer interaction~\cite{hassan2021populating}, virtual reality~\cite{yang2022hybridtrak}, and sports analysis~\cite{ingwersen2023sportspose}. However, HPE faces two fundamental challenges: 1) the difficulty of achieving expressive and detailed body reconstruction (e.g., 17-joint representations are too sparse to capture fine-grained motion), and 2) severe self-occlusion issues commonly encountered during flexible human body movements. These challenges are often entangled, making it extremely difficult to design a model that is both expressive and robust to occlusion.

Naturally, accurate 2D coordinate estimation provides a strong foundation for obtaining satisfactory 3D keypoint locations using off-the-shelf 3D lifting models \cite{chen2021anatomy, zhang2022mixste, gong2023diffpose, shan2023diffusion, yu2023gla}, which are specifically designed to map 2D poses into 3D space. However, the inherently limited expressiveness of sparse 2D inputs (e.g., 17 joints) creates a severe information bottleneck, making the 3D lifting process highly ambiguous and prone to structural collapse, especially under occlusion. As illustrated in Figure~\ref{fig:varpose_frameworks}(a), relying solely on 17-joint inputs often leads to distorted 3D reconstructions, as the model lacks sufficient localized structural cues to resolve depth ambiguity. To fundamentally overcome this limitation, densifying 2D poses offers a natural solution, as dense joints provide a more expressive representation for ``right elbow'', as an example, leading to more focused attention and reduced depth ambiguity, as shown in Figure~\ref{fig:varpose_frameworks}(b). Our core motivation is to enrich 2D pose representations with dense intermediate joints to provide richer context and stronger geometric constraints before the 3D lifting stage. As illustrated in Figure~\ref{fig:varpose_frameworks}(c), existing work like HiPART \cite{zheng2025hipart} studies the hierarchical relation between poses of different granularities (e.g., 48-joint and 96-joint) by learning independent sparse and dense codebooks and performing post-hoc skeleton-aware alignment. Nevertheless, we argue that achieving semantic alignment after separately learning codebooks for different granularities inherently suffers from semantic misalignment, as well as limitations in flexibility and scalability. This design still relies on next-token generation, which primarily captures local joint topology while neglecting the global structural coherence of the pose.

Meanwhile, in natural image generation, Visual AutoRegressive Modeling (VAR) \cite{tian2024visual} has popularized a \textbf{next-scale prediction} paradigm that prevails in high-fidelity synthesis. VAR learns a unified multi-scale codebook and synthesizes images in a coarse-to-fine manner, enabling faithful high-definition generation. We ask: \hl{\textit{Can such a unified representation be exploited for human poses across multiple granularities?}} We answer in the affirmative. Drawing an analogy between image resolution and pose granularity (e.g., low-resolution $\leftrightarrow$ sparse pose; high-resolution $\leftrightarrow$ dense pose), we argue that the ``next-scale prediction'' paradigm from VAR can be effectively transposed to the structured domain of human skeletons. To this end, we introduce \textbf{VARPose}, a novel framework that pioneers \textbf{next-granularity pose prediction} as shown in Figure~\ref{fig:varpose_frameworks}(d). VARPose materializes this concept by reinterpreting VAR's fundamental unit of ``scale'' as skeletal ``granularity'', enabling unified and hierarchical generation of human poses from coarse to fine. 

Our VARPose framework consists of two core components: a Granularity-agnostic Pose Tokenizer (GPT) and a Unified Skeletal Autoregressive model (UniSkelar). First, GPT leverages a single hybrid codebook with residual quantization to encode poses of varying densities into a unified, multi-scale discrete representation. Then, UniSkelar learns to predict this token sequence in a coarse-to-fine manner, conditioned on the sparsest pose. This two-stage process ensures that both the tokenization and generation phases are optimized for hierarchical, granularity-agnostic pose synthesis. By modeling global skeletal dependencies rather than independent joint tokens, VARPose inherently demonstrates remarkable robustness against imprecise initial poses. As validated on challenging real-world benchmarks such as 3DPW-Occ~\cite{zhang2020object}, our framework effectively rectifies detection-induced errors and mitigates structural collapses even under severe noise or heavy self-occlusion. 

Furthermore, VARPose not only learns fine-grained alignment and relationships between different pose granularities but can also be easily extended to novel pose synthesis. While our generative densification pipeline produces high-fidelity dense poses, integrating them into downstream 3D lifting tasks requires careful design to preserve the integrity of existing backbones with minimal architectural modification. To address this, we propose two plug-in strategies to inject densified skeletal information for lifting 2D keypoints to accurate 3D poses, as illustrated in Figure~\ref{fig:varpose_frameworks}(e). The resulting 3D poses can further drive human surface models such as SMPL via solvers like SMPL-IKS~\cite{10.1007/s11263-025-02574-5}. We also conduct extensive ablation studies, showing the effectiveness of densification and the reason why it benefits downstream 3D tasks. In summary, our main contributions are as follows: 
\begin{itemize}[noitemsep, nosep, left=0pt]
    \item We propose a \textbf{new problem setting}, granularity-agnostic 2D human pose densification, that supports novel pose generation. Crucially, our cross-granularity codebook representation generalizes remarkably well. Under this decoupled representation-projection paradigm, we adapt to entirely new pose granularities by simply retraining the decoder with a frozen codebook.
    \item We are the first to propose \textbf{VARPose}, a next-granularity autoregressive framework for 2D human pose densification that extends VAR's ``next-scale'' prediction by reinterpreting image scale as skeletal joint density. Unlike VAR, which generates images only from the final scale, our UniSkelar yields meaningful intermediate pose granularities, ensuring structural integrity from the sparsest to the densest skeleton. By enforcing global structure consistency across granularities, VARPose naturally achieves SOTA in the densification task. 
    \item We validate two \textbf{plug-in} strategies for injecting dense poses into downstream tasks: joint-wise concatenation with full retraining and cross-attention fusion with fine-tuning. Ablations demonstrate a sensible performance trade-off between them.
\end{itemize}
\section{Related Work}
\subsection{3D Human Pose Estimation}
The dominant paradigm of 3D Human Pose Estimation is two-stage lifting: first, a 2D pose detector extracts keypoint coordinates from images or videos \cite{cao2017realtime, chen2018cascaded}, and then a 3D lifting model regresses them into 3D space \cite{chen2021anatomy, gong2023diffpose, shan2023diffusion}. The performance of this pipeline is therefore heavily reliant on the quality of the 2D pose input. To improve robustness, mainstream research has explored several directions for utilizing different sources of information. VideoPose3D \cite{pavllo20193d} uses temporal convolutions over a sequence of 2D poses to enforce motion continuity, smoothing out inaccuracies and inferring occluded joints. This concept is significantly advanced by Transformer-based architectures such as PoseFormer \cite{zheng20213d}, MHFormer \cite{li2022mhformer}, and MixSTE \cite{zhang2022mixste}, which excel at modeling long-range spatio-temporal dependencies for more robust and coherent pose sequences. More recently, generative approaches, particularly diffusion models \cite{ci2023gfpose, shan2023diffusion, wang2024text}, have gained prominence. DiffPose \cite{gong2023diffpose} frames 3D lifting as a denoising process. It starts with a noisy 3D pose and iteratively refines it conditioned on the 2D input, leading to high-quality and plausible results. While powerful, their performance is fundamentally limited by the sparse 2D pose input, which can be an irreversible bottleneck in cases of severe occlusion or ambiguity. This insight has spurred a new research direction: generatively enhancing the 2D pose itself before the 3D lifting stage. The pioneering work, HiPART \cite{zheng2025hipart}, demonstrates the feasibility of generating dense hierarchical poses (e.g., 48 and 96 joints) from a sparse input using an autoregressive model. However, its framework is inherently limited to a closed set of predefined granularities seen during training, and its ``next-token prediction'' paradigm captures local joint relationships while neglecting global structural coherence, potentially producing anatomically implausible poses. In contrast, our work proposes a fundamentally different generative framework to overcome these limitations.

\subsection{Autoregressive Modeling}  
``Next-token prediction'' is the core mechanism of autoregressive models, a paradigm popularized in Natural Language Processing (NLP) by models such as the Generative Pre-trained Transformer (GPT) \cite{radford2018improving, touvron2023llama}, which predict the next discrete token in a sequence given all preceding elements. The success of this paradigm has extended to visual tasks such as image and video generation \cite{chen2020generative, he2025neighboring}, where visual data are tokenized and flattened into a 1D sequence of patches or tokens and predicted one after another. In the realm of pose estimation, HiPART also employs the next-token paradigm to generate denser skeletons. However, its sequential, raster-scan nature of next-token prediction is ill-suited for spatial data like images, as it disregards the crucial 2D spatial relationships between tokens. To address this, VAR \cite{tian2024visual} introduces the revolutionary ``next-scale prediction'' paradigm, which generates all tokens for a given spatial resolution in parallel, proceeding from coarse to fine scales. Owing to its effectiveness, next-scale thinking has been rapidly adopted for various high-fidelity visual generation tasks, including text-to-image generation \cite{tang2024hart}, all-in-one image restoration \cite{wang2025navigating}, image super-resolution \cite{pmlr-v267-qu25h}, and high-resolution image synthesis \cite{han2025infinity}. \textbf{Yet, the potential of ``next-scale prediction'' remains unexplored for structured, non-image data like human skeletons.} We argue that there is a natural analogy between ``scale'' in images and ``granularity'' in human poses. A low-resolution image corresponds to a sparse skeleton, while a high-resolution image corresponds to a dense one. This paper is the first to transpose VAR from image scales to pose granularities, allowing us to overcome the global modeling limitations of previous methods and unlock open-set, granularity-agnostic pose generation.

\subsection{Human Mesh Recovery from Poses}
Human Mesh Recovery (HMR) extends 3D HPE by reconstructing the full 3D surface of the human body, typically represented by parametric body models such as SMPL \cite{10.1145/2816795.2818013}. Many methods regress mesh vertices or body-model parameters directly from RGB images \cite{kanazawa2018end, kolotouros2019learning, moon2020i2l}. However, their predictions can be affected by clothing, backgrounds, and appearance-domain variations. An alternative is to use estimated human poses as intermediate representations, reducing dependence on image appearance and decoupling pose estimation from surface reconstruction.
Pose-conditioned HMR can be formulated in different ways. Pose2Mesh~\cite{choi2020pose2mesh}, for example, directly maps a 2D pose to 3D mesh vertices using a coarse-to-fine GraphCNN that exploits human articulation and mesh topology. SMPL-IKS~\cite{10.1007/s11263-025-02574-5}, by contrast, recovers SMPL shape and pose parameters from a 3D skeleton through a mixed analytical-neural inverse kinematics solver, explicitly addressing shape mismatch, error accumulation, and rotation ambiguity.
Sparse 2D skeletons provide insufficient geometric constraints, often leading to imprecise 3D joint estimates, particularly for limb rotations, which remain largely underdetermined during subsequent mesh recovery. Existing pipelines may therefore be inherently bottlenecked by the limited structural information available at the lifting stage. In this work, VARPose, together with a plug-in lifting strategy, eliminates this bottleneck through a densify-then-lift paradigm. To clearly isolate and quantify the downstream gains, we deliberately adopt SMPL-IKS, a simple, lightweight inverse-kinematics-based solver, as the mesh recovery module.
\begin{figure*}[t]
  \centering
  \includegraphics[width=\textwidth]{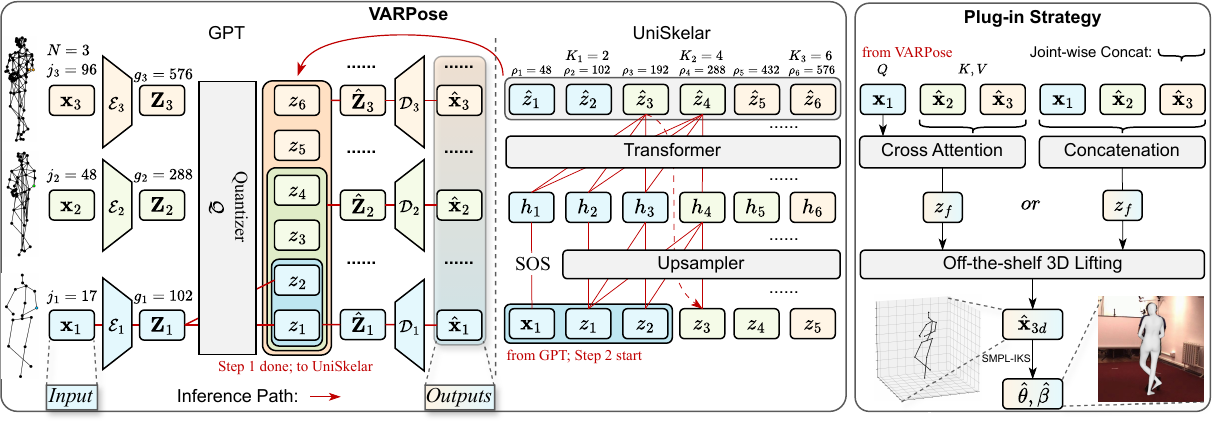}
  \caption{Architecture of VARPose and the plug-in strategy pipeline. During inference, given the sparsest input pose, VARPose can generate poses from a 17-joint input at flexible target granularities. Our plug-in strategies, cross-attention fusion and joint-wise concatenation, enable downstream models to utilize VARPose's dense outputs as a shape prior.}
  \Description{Architecture of VARPose and the plug-in strategy pipeline. During inference, given the sparsest input pose, VARPose can generate poses from a 17-joint input at flexible target granularities. Our plug-in strategies, cross-attention fusion and joint-wise concatenation, enable downstream models to utilize VARPose's dense outputs as a shape prior.}
  \label{fig:onecol}
\end{figure*}
\section{Methodology}
\subsection{Preliminaries: Next-scale Prediction}
\label{sub:preliminary}
VAR reconceptualizes autoregressive learning on images by shifting from ``next-token prediction'' to a ``next-scale prediction'' strategy. Considering an input fine-grained feature map $f\in\mathbb{R}^{H\times W\times C}$ for an image, a multi-scale quantizer with a shared codebook of size $V$ tokenizes it into a sequence of $K$ discrete token maps, denoted as $R=(r_1, r_2, \dots, r_K)$, where each map $r_k\in[V]^{h_k\times w_k}$. The autoregressive likelihood function is defined as:
\begin{equation}
    p(r_1, r_2, \dots, r_K) = \prod_{k=1}^K p(r_k|r_{< k}, c),
    \label{eq:likelihood}
\end{equation}
where $r_{< k}$ denotes the conditional prefix used to predict the token map $r_k$, and $c$ represents the conditioning information. At each autoregressive step $k$, the model generates the $h_k\times w_k$ tokens of $r_k$ in parallel, leveraging its prefix $r_{< k}$ and the conditions $c$.

\subsection{Reformulation and Method Overview}
Considering the strong correlation between image generation and pose densification, we introduce VAR into the pose estimation domain. Given multi-scale 2D poses $\mathbf{X}=\{\mathbf{x}_n\}_{n=1}^N$, encoders $\{\mathcal{E}_n\}_{n=1}^N$ project each $\mathbf{x}_n$ into features $\mathbf{Z}_n$ at granularity $g_n$. A Granularity-agnostic Pose Tokenizer (GPT) with codebook $\mathcal{C}$ of size $V$ tokenizes all features through multi-level residual quantization into a token sequence $\mathbf{R}=\{r_k\}_{k=1}^K$ and quantized features $\{z_k\}_{k=1}^K$, where each $r_k$ possesses a one-dimensional quantization density $\rho_k$ that increases with level $k$, replacing the 2D spatial resolution in the original VAR (\textbf{Step 1}). The autoregressive likelihood is $p(\mathbf{R}) = \prod_{k=1}^K p(r_k|r_{< k}, \mathbf{x}_1)$, where $r_{< k}$ is the conditional prefix and the sparsest pose $\mathbf{x}_1$ serves as the start-of-sequence (SOS) (\textbf{Step 2}). The predicted tokens $\hat{\mathbf{R}}$ are decoded by pre-trained decoders $\{\mathcal{D}_n\}_{n=1}^N$ into 2D poses $\hat{\mathbf{X}}$ of flexible joint granularity. The frozen codebook decouples representation from specific joint topologies, enabling the synthesis of novel granularity poses. For downstream 3D lifting, a Plug-in Strategy fuses multi-scale poses via cross-attention fusion or joint-wise concatenation into $z_f$, followed by the off-the-shelf 3D lifting model and SMPL-IKS to produce a 3D skeleton $\hat{\mathbf{x}}_{3d}$ and SMPL parameters $(\hat{\theta}, \hat{\beta})$. See Table~\ref{tab:notation} for a summary of the notation and Figure~\ref{fig:onecol} for the full framework.

During training, GPT (Sec.~\ref{sub:GPT}) quantizes the encoded features into $\mathbf{R}$ and $\{z_k\}_{k=1}^K$, and reconstructs the input poses through $\{\mathcal{D}_n\}_{n=1}^N$ to learn a unified codebook. UniSkelar (Sec.~\ref{sub:UniSkelar}) is then trained to autoregressively predict $\hat{\mathbf{R}}$ over $\{h_k\}$ via a Transformer, conditioned on the SOS $\mathbf{x}_1$, with an Upsampler aligning $\hat{z}_k$ to target quantization densities at each level.
\begin{table}[t]
\centering
\small
\caption{Summary of notations used in the VARPose framework.}
\label{tab:notation}
\setlength{\tabcolsep}{2pt}
\resizebox{0.96\linewidth}{!}{
\begin{tabular}{llllllll}
\hline
Sym. & Descript.& Dom. & Sym. & Descript. & Dom. & Sym. & Descript. \\
\hline
 $  N  $     & num poses          &  $  \mathbb{Z}^+  $  &
 $  \rho_k  $  & quant density   &  $  \mathbb{Z}^+  $  &
 $\mathcal{Q}$ & quantizer\\  

 $  j_n  $   & num joints         &  $  \mathbb{Z}^+  $  &
 $  z_k  $   & quant feature    &  $  \mathbb{R}^{\rho_k\times D}  $  &  
 $  \mathcal{E}_n  $  & $n$th encoder          \\
 
$  \mathbf{x}_n  $  & 2D pose    &  $  \mathbb{R}^{j_n\times 2}  $  &
 $  h_k  $   & AR input          &  $  \mathbb{R}^{\rho_k\times D}  $  &  
$ \mathcal{D}_n  $  & $n$th decoder         \\
 
 $  g_n  $   & granularity       &  $  \mathbb{Z}^+  $  &
  $  K_n  $     & quant steps       &  $  \mathbb{Z}^+  $  &
 SOS & AR start\\ 
 
$  D  $     & feature dim          &  $  \mathbb{Z}^+  $  &
$  r_k  $   & token indices     &  $  \mathbb{Z}^{\rho_k\times 1}  $  &
$  \phi  $  & MLP ResNet      \\

 $  \mathbf{Z}_n  $  & 2D feature   &  $  \mathbb{R}^{g_n\times D}  $  &
 $  \varphi_{k,n}  $  & residual     &  $  \mathbb{R}^{\rho_{K_n}\times D}  $  & 
 $  \mathcal{D}  $    & decoders    \\
\hline
\end{tabular}
}
\end{table}
\subsection{GPT: Granularity-agnostic Pose Tokenizer}
\label{sub:GPT}
The use of pose-specific codebooks suffers from poor scalability and hinders the modeling of relationships across diverse poses. However, quantizing multi-granularity features with a single codebook to obtain the unified codebook is non-trivial. To overcome these limitations, our GPT instead adopts a unified codebook architecture to sufficiently consider the correlation between poses of different joint numbers.
 
\paragraph{Quantization and Reconstruction} 
The encoder $\mathcal{E}_n$ first projects the input pose $\mathbf{x}_n$ into features $\mathbf{Z}_n$ with granularity $g_n$ and dimension $D$. Given a maximum of $K$ quantization steps, the number of quantization steps $K_n$ assigned to $\mathbf{Z}_n$ is determined by:
\begin{equation}
K_n=\max\limits_{i \in \{1, \dots,K\}}\{\rho_i \leq g_n\}.
\end{equation}
Prior to the quantization of $\mathbf{Z}_n$, to facilitate codebook learning of unified features, we update $\mathbf{Z}_n$ by subtracting the residuals derived from the quantized features of previous levels $z_{\leq K_{n-1}}$ via an interpolation mechanism and an MLP ResNet $\phi(x)=\frac{x+\text{MLP}(x)}{2}$ with $4D$ hidden units and ReLU activation. This process is formulated as:
\begin{equation}
    \varphi_{k,n} = \phi(\text{interpolate}(z_k, \rho_{K_n})); \quad \mathbf{Z}_n \leftarrow \mathbf{Z}_n - \sum_{k=1}^{K_{n-1}}\varphi_{k,n},
\end{equation}
noting that $\rho_{K_n}$ typically aligns with $g_n$. Subsequently, the features $\mathbf{Z}_n$ undergo interpolation and quantization at densities $\{\rho_k\}$, where $K_{n-1}<k\leq K_n$, to obtain the residual quantized feature $z_k$ and the residual token $r_k$. The feature map $\mathbf{Z}_n$ is then updated by subtracting the reconstructed residual of the current level. Formally, for each level $k$:
\begin{equation}
(z_k, r_k)=\mathcal{Q}(\text{interpolate}(\mathbf{Z}_n, \rho_k)); \quad \mathbf{Z}_n \leftarrow \mathbf{Z}_n - \varphi_{k,n}. \label{eq:quantization_per_level}
\end{equation}

After acquiring all residual quantized features, we reconstruct $\hat{\mathbf{Z}}_n$ by aggregating these details, a strategy analogous to that used in image restoration. The reconstruction is formulated as:
\begin{equation}
     \hat{\mathbf{Z}}_n = \sum_{k=1}^{K_n} \varphi_{k,n}.
     \label{eq:fusing_latent_features}
\end{equation}
Finally, $\hat{\mathbf{Z}}_n$ is fed into the corresponding decoder $\mathcal{D}_n$ to yield the reconstructed pose $\hat{\mathbf{x}}_n$. In summary, the overview of the training procedure for GPT is shown in Algorithm~\ref{alg:GPT_forward}.

\paragraph{Unified Codebook for Novel Pose Generation}
For novel poses $\{\mathbf{x}^{\dagger}_{m}\}_{m=1}^{M}$ unseen during initial training, VARPose reconstructs them from quantized features by retraining only lightweight decoders $\{\mathcal{D}^{\dagger}_{m}\}_{m=1}^{M}$ without updating the codebook. This is feasible because the unified codebook learns granularity-agnostic pose representations independent of specific joint layouts. Given each unseen granularity, the lightweight decoder serves only as a projection head that maps shared latent features to the corresponding 2D coordinates. Separation between representation and projection significantly reduces the generalization burden for novel poses.

However, the unseen granularity $g_m^\dagger$ of novel poses still presents a challenge for direct alignment with the fixed quantization densities $\{\rho_k\}$. To address this, we adopt a strategy to determine the appropriate quantization steps $K_m^\dagger$, formally defined as:
\begin{equation}
K_m^\dagger = \begin{cases}
1,& g^{\dagger}_{m} \leq \rho_1\\
k,& \rho_{k-1} < g^{\dagger}_{m} \leq \rho_{k} \\
K,& g^{\dagger}_{m} > \rho_K\\
\end{cases}.
\end{equation}
Since the granularity of the reconstructed features from Eq.~(\ref{eq:fusing_latent_features}) may not align with the target $g_m^\dagger$, we introduce a final interpolation step to adjust the resolution. This is formulated as:
\begin{equation}
\hat{\mathbf{Z}}^{\dagger}_{m} = \text{interpolate}(\hat{\mathbf{Z}}_m, g^{\dagger}_{m}),
\end{equation}
where $\hat{\mathbf{Z}}_m$ represents the aggregated features obtained via Eq.~(\ref{eq:fusing_latent_features}).

\begin{algorithm}[t]
\caption{GPT Training}
\label{alg:GPT_forward}
\begin{algorithmic}[1]
\Input{2D Poses $\{\mathbf{x}_n\}_{n=1}^N$}
\Hyperparams{Quant Steps $\{K_n\}_{n=1}^{N}$, Quant Densities $\{\rho_k\}_{k=1}^K$, Target Pose Granularity $\{g_n\}_{n=1}^N$}
\For{$n = 1, \dots, N$}
    \State $\mathbf{Z}_n = \mathcal{E}_n(\mathbf{x}_n)$; $\hat{\mathbf{Z}}_n = 0$; 
    \For{$k = 1, \dots, K_n$}
        \If{$n = 1$ or $k > K_{n-1}$}
            \State $(z_k, r_k) = \mathcal{Q}(\text{interpolate}(\mathbf{Z}_n, \rho_k))$
        \EndIf
        \State $\varphi_{k,n} = \phi(\text{interpolate}(z_k, \rho_{K_n}))$
        \State $\mathbf{Z}_n = \mathbf{Z}_n - \varphi_{k,n}$; $\hat{\mathbf{Z}}_n = \hat{\mathbf{Z}}_n +\varphi_{k,n}$
    \EndFor
    \State $\hat{\mathbf{x}}_n = \mathcal{D}_n(\hat{\mathbf{Z}}_n)$
\EndFor
\State $\mathbf{R} = (r_1,\dots,r_K)$; $\hat{\mathbf{X}} = (\hat{\mathbf{x}}_1, \dots, \hat{\mathbf{x}}_N)$
\State\Return{Residual token: $\mathbf{R}$, Reconstructions: $\hat{\mathbf{X}}$}
\end{algorithmic}
\end{algorithm}

\subsection{UniSkelar: Unified Skeletal AR Generation}
\label{sub:UniSkelar}
We design UniSkelar to densify the sparsest pose $\mathbf{x}_1$. Its next-density prediction paradigm leverages full skeletal poses as context, utilizing anatomical information more effectively than local token-based methods. Combined with our granularity-agnostic codebook, this simplifies densification by generating general pose features in the quantized space. The overall framework is illustrated in Figure~\ref{fig:onecol}.

The model inputs are denoted as $\{h_k\}_{k=1}^K$. The first input $h_1$ is initialized via an MLP-Mixer~\cite{tolstikhin2021mlp} transformation of $\mathbf{x}_1$. Subsequent inputs $h_k$ ($k > 1$) are derived by aggregating latent features $z_{<k}$ and upsampling them to the target scale $\rho_k$ and the target dimension $D$:
\begin{equation}
      \mathbf{Z}_{fusion_k}=\sum_{i=1}^{k-1} \varphi_{i,N};\quad h_{k} = \text{Linear}(\text{interpolate}(\mathbf{Z}_{fusion_k}, \rho_k)),
\end{equation}
where $\mathbf{Z}_{fusion_k}$ represents the fusion result at the finest granularity according to Eq.~(\ref{eq:fusing_latent_features}). This module serves as the ``\textbf{Upsampler}''.

The model employs a Transformer with Rotary Position Embeddings (RoPE)~\cite{su2024roformer} to better encode relative positional relationships among skeletal tokens. During inference, the model autoregressively predicts logits $l$ over $\mathbf{R}$. Then, sampling yields indices $\hat{\mathbf{R}}$, which are projected to generate the final pose sequence $\hat{\mathbf{X}}$. We utilize a causal mask and the KV-Cache mechanism for efficient generation. The inference pipeline is detailed in Algorithm~\ref{alg:inference}.

\subsection{Loss Functions}
For the GPT model, the total loss is a weighted combination of two components, the reconstruction loss $\mathcal{L}_{Recon}$ and the quantization loss $\mathcal{L}_{VQ}$, described as:
\begin{equation}
\begin{gathered}
    \mathcal{L}_{\text{GPT}} = \lambda \mathcal{L}_{\text{Recon}} + \mathcal{L}_{\text{VQ}} \\
    \mathcal{L}_{\text{Recon}} = \sum_n \|\mathbf{x}_n - \hat{\mathbf{x}}_n\|^2 \\
    \mathcal{L}_{\text{VQ}} = \sum_n \left( \| \operatorname{sg}(\mathbf{Z}_n) - \hat{\mathbf{Z}}_n \|^2 + \| \mathbf{Z}_n - \operatorname{sg}(\hat{\mathbf{Z}}_n) \|^2 \right),
\end{gathered}
\end{equation}
where $\lambda$, empirically set to 200, is a hyperparameter balancing the two terms and $\operatorname{sg}(\cdot)$ denotes the stop-gradient operator. Codebook updates are performed using the straight-through estimator in conjunction with an Exponential Moving Average (EMA) strategy.

For the UniSkelar model, which outputs logits $l$ over the target variable $\mathbf{R}$, we utilize cross-entropy loss as:
\begin{equation}
    \mathcal{L}_{\text{UniSkelar}} = -\sum_{k} \log \frac{\exp(l_{k,r_k})}{\sum_j \exp(l_{k,j})},
\end{equation}
where $r_k$ denotes the ground-truth indices, and $l_{k,j}$ is the predicted logit for indices $j$, both with quantization density $\rho_k$.

For the unseen pose decoder, we employ solely the reconstruction loss $\mathcal{L}_{Recon}$, utilizing the reconstructed features $\hat{\mathbf{Z}}_m^\dagger$ derived from the frozen GPT model.

\begin{algorithm}[t]
\caption{Inference}
\label{alg:inference}
\begin{algorithmic}[1]
\Input{Sparsest 2D Pose $\mathbf{x}_1$}
\Hyperparams{Quant Densities $\{\rho_k\}_{k=1}^K$, Codebook $\mathcal{C}$, Quantizer $\mathcal{Q}$, Decoder $\mathcal{D}$}
\State $K_1=\max\limits_{i=1, \dots,K}\{\rho_i \leq g_1\}$
\State $h_1 = \text{MlpMixer}(\mathbf{x}_1)$; $(\{\hat{z}_k\}_{k=1}^{K_1}, \{\hat{r}_k\}_{k=1}^{K_1}) \leftarrow \mathcal{Q}(\mathbf{x}_1)$

\For{$k=1, \dots, K$}
    \State $l_k \leftarrow \text{GetLogits}(\text{Transformers}(h_k))$
    \If{$k > K_1$}
        \State $\hat{r}_k\sim l_k$; $\hat{z}_k = \mathcal{C}(\hat{r}_k)$
    \EndIf
    \State $h_{k+1}\leftarrow\text{Upsampler}(\hat{z}_{<k+1})$
\EndFor
\State $\hat{\mathbf{X}} = \mathcal{D}(\{\hat{z}_k\}_{k=1}^{K})$
\State \Return Predicted Pose $\hat{\mathbf{X}}$
\end{algorithmic}
\end{algorithm}

\subsection{Plug-in Strategy in 3D Lifting}
\label{sec:lifting_plug_in_strategy}
Off-the-shelf 3D lifting models cannot directly utilize VARPose's dense 2D poses. As illustrated in Figure~\ref{fig:onecol}, we adopt two distinct strategies to inject this information:

\begin{itemize}[noitemsep, nosep, left=0pt]
    \item \textbf{Joint-wise Feature Concatenation}: We concatenate all features in a joint-wise manner to form an extended input vector, requiring a dedicated regression head and \textbf{retraining} of the model.
    
    \item \textbf{Cross-Attention Fusion}: We utilize the original 17-joint features as queries ($\mathbf{Q}$) and dense features as keys ($\mathbf{K}$) and values ($\mathbf{V}$) to enrich the representation:
    \begin{equation}
        \mathbf{X}_{\text{input}} = \text{LayerNorm}\left(\mathbf{X} + \text{softmax}\left(\frac{\mathbf{Q}\mathbf{K}^\top}{\sqrt{D}}\right)\mathbf{V}\right).
    \end{equation} 
    This design allows for flexible \textbf{fine-tuning} of the lifting model.
\end{itemize}
\section{Experiments}
\begin{table*}[t]
  \centering
  \caption{Quantitative comparison with state-of-the-art methods on Human3.6M using MPJPE. Results are grouped by the source of 2D poses: SH-detected poses~\cite{newell2016stacked} and ground-truth 2D poses. $S$ denotes the number of hypotheses, and $f$ represents the number of input frames for temporal-based models. The best and second-best results are highlighted in \textbf{bold} and \underline{underlined}.}
  \label{tab:human3.6m_comparison}
  \resizebox{\textwidth}{!}{
  \begin{tabular}{l|ccccccccccccccc|c}
    \toprule
    Method & Dir. & Disc. & Eat & Greet & Phone & Photo & Pose & Purch. & Sit & SitD. & Smoke & Wait & WalkD. & Walk & WalkT. & Avg. \\
    \midrule
    Learning \cite{fang2018learning} & 50.1 & 54.3 & 57.0 & 57.1 & 66.6 & 73.3 & 53.4 & 55.7 & 72.8 & 88.6 & 60.3 & 57.7 & 62.7 & 47.5 & 50.6 & 60.4 \\
    SemGCN \cite{zhao2019semantic} & 47.3 & 60.7 & 51.4 & 60.5 & 61.1 & 49.9 & 47.3 & 68.1 & 86.2 & 55.0 & 67.8 & 61.0 & 42.1 & 60.6 & 45.3 & 57.6 \\
    Monocular \cite{xu2021monocular} & 47.1 & 52.8 & 54.2 & 54.9 & 63.8 & 72.5 & 51.7 & 54.3 & 70.9 & 85.0 & 58.7 & 54.9 & 59.7 & 43.8 & 47.1 & 58.1 \\
    Graformer \cite{zhao2022graformer} & 49.3 & 53.9 & 54.1 & 55.0 & 63.0 & 69.8 & 51.1 & 53.3 & 69.4 & 90.0 & 58.0 & 55.2 & 60.3 & 47.4 & 50.6 & 58.7 \\
    Lifting by Image \cite{zhou2024lifting} & 48.3 & 51.5 & 46.1 & \underline{48.5} & 53.7 & 42.8 & 47.3 & 59.9 & 71.0 & \textbf{51.6} & 52.7 & 46.1 & 39.8 & 53.0 & 43.9 & 51.0 \\
    GFPose \cite{ci2023gfpose}$(S\!=\!200)$ & \underline{31.7} & \textbf{35.4} & \underline{31.7} & \textbf{32.3} & \underline{36.4} & \underline{42.4} & \underline{32.7} & \underline{31.5} & \textbf{41.2} & 52.7 & \underline{36.5} & \underline{34.0} & \textbf{36.2} & \underline{29.5} & \underline{30.2} & \underline{35.6} \\
    \midrule
    GFPose$(S\!=\!200)$ + VARPose (Ours)  & \textbf{29.2}\textcolor{gray}{$\Delta$2.5} & \underline{35.9}\textcolor{gray}{$\Delta$-0.5} & \textbf{30.6}\textcolor{gray}{$\Delta$1.1} & \textbf{32.3}\textcolor{gray}{$\Delta$0.0} & \textbf{34.7}\textcolor{gray}{$\Delta$1.7} & \textbf{41.7}\textcolor{gray}{$\Delta$0.7} & \textbf{31.7}\textcolor{gray}{$\Delta$1.0} & \textbf{30.3}\textcolor{gray}{$\Delta$1.2} & \underline{43.6}\textcolor{gray}{$\Delta$-2.4} & \underline{51.8} \textcolor{gray}{$\Delta$0.9} & \textbf{35.5}\textcolor{gray}{$\Delta$1.0} & \textbf{33.6}\textcolor{gray}{$\Delta$0.4} & \underline{36.9}\textcolor{gray}{$\Delta$-0.7} & \textbf{27.1}\textcolor{gray}{$\Delta$2.4} & \textbf{29.8}\textcolor{gray}{$\Delta$0.4} & \textbf{35.0}\textcolor{gray}{$\Delta$0.6} \\
    \midrule\midrule

    SemGCN \cite{zhao2019semantic} & 37.8 & 49.4 & 37.6 & 40.9 & 45.1 & 41.4 & 40.1 & 48.3 & 50.1 & 42.2 & 53.5 & 44.3 & 40.5 & 47.3 & 39.0 & 43.8 \\
    Pose2Mesh \cite{choi2020pose2mesh} & 38.1 & 41.7 & 38.3 & 37.5 & 39.2 & 45.4 & 37.5 & 36.2 & 45.7 & 50.1 & 39.8 & 39.2 & 40.2 & 35.2 & 37.6 & 40.1 \\
    HGN \cite{li2021hierarchical} & 35.4 & 40.2 & 31.1 & 38.2 & 38.3 & 41.1 & 36.1 & 32.7 & 42.1 & 48.4 & 37.1 & 36.9 & 37.1 & 30.5 & 32.4 & 37.2 \\
    PoseFormer \cite{zheng20213d}$(f\!=\!81)$ & 30.0 & 33.6 & 29.9 & 31.0 & 30.2 & 33.3 & 34.8 & 31.4 & 37.8 & 38.6 & 31.7 & 31.5 & 29.0 & 23.3 & 23.1 & 31.3 \\
    MHFormer \cite{li2022mhformer}$(f\!=\!351)$ & 27.7 & 32.1 & 29.1 & 28.9 & 30.0 & 33.9 & 33.0 & 31.2 & 37.0 & 39.3 & 30.0 & 31.0 & 29.4 & 22.2 & 23.0 & 30.5 \\
    MixSTE \cite{zhang2022mixste}$(f\!=\!81)$ & \underline{25.6} & \underline{27.8} & \underline{24.5} & \underline{25.7} & \underline{24.9} & 29.9 & 28.6 & \underline{27.4} & \underline{29.9} & \underline{29.0} & \underline{26.1} & \underline{25.0} & \underline{25.2} & \underline{18.7} & \underline{19.9} & \underline{25.9} \\
    DiffPose \cite{gong2023diffpose} & 28.8 & 32.7 & 27.8 & 30.9 & 32.8 & 38.9 & 32.2 & 28.3 & 33.3 & 41.0 & 31.0 & 32.1 & 31.5 & 25.9 & 27.5 & 31.6 \\
    Lifting by Image \cite{zhou2024lifting}& 29.5 & 30.1 & 25.0 & 29.0 & 28.5& \underline{28.6} & \underline{26.9} & 30.5 & 31.1 & \textbf{27.7} & 32.4 & 27.7 & \textbf{24.8} & 30.0 & 25.9 & 28.6 \\
    HiPART \cite{zheng2025hipart} & 30.4 & 29.7 & 26.3 & 27.2 & 28.7 & 29.1 & 28.2 & 29.2 & 30.9 & 33.1 & 29.6 & 26.2 & 27.2 & 21.9 & 26.2 & 28.3 \\
    \midrule
    MixSTE $(f\!=\!81)$ + VARPose(Ours) & \textbf{24.1}\textcolor{gray}{$\Delta$1.5} &\textbf{26.3}\textcolor{gray}{$\Delta$1.5} &\textbf{22.5}\textcolor{gray}{$\Delta$2.0} & \textbf{23.1}\textcolor{gray}{$\Delta$2.6} & \textbf{24.3}\textcolor{gray}{$\Delta$0.6} & \textbf{27.9}\textcolor{gray}{$\Delta$2.0} & \textbf{26.7}\textcolor{gray}{$\Delta$1.9} & \textbf{25.6}\textcolor{gray}{$\Delta$1.8} & \textbf{29.7}\textcolor{gray}{$\Delta$0.2} & 32.0\textcolor{gray}{$\Delta$-3.0} & \textbf{25.5}\textcolor{gray}{$\Delta$0.6} & \textbf{24.3}\textcolor{gray}{$\Delta$0.7} & 25.4\textcolor{gray}{$\Delta$-0.2} &\textbf{17.7}\textcolor{gray}{$\Delta$1.0} & \textbf{19.6}\textcolor{gray}{$\Delta$0.3} & \textbf{25.0}\textcolor{gray}{$\Delta$0.9} \\
    \bottomrule
  \end{tabular}
  }
\end{table*}
\subsection{Datasets and Evaluation Metrics}
\paragraph{Datasets} \textbf{Human3.6M}~\cite{ionescu2013human3} is one of the most widely used and largest benchmark datasets for the field of 3D human pose estimation. The dataset comprises 3.6 million video frames of 15 daily activities performed by 11 actors (6 males and 5 females) with corresponding 3D annotations. Following the convention \cite{zhang2022mixste, ci2023gfpose}, we train on subjects 1, 5, 6, 7, 8 and evaluate on subjects 9 and 11. \textbf{MPI-INF-3DHP}~\cite{mehta2017monocular} aims to bridge the gap between laboratory environments and in-the-wild applications, encompassing three types of scenes: green-screen scenes, non-green-screen scenes, and outdoor scenes. The dataset records motion sequences of 8 actors performing various activities using 14 cameras, with the training set containing 8 activities and the test set containing 7 activities. \textbf{3DPW}~\cite{von2018recovering} is the first in-the-wild dataset to provide accurate 3D poses taken from a moving phone camera. It contains 60 video sequences across various activities of 7 subjects. Following the standard protocol~\cite{zheng2025hipart, wang2024text}, we also evaluate our method on this dataset to measure robustness and generalization. Furthermore, we report results on 3DPW-Occ~\cite{zhang2020object}, a subset of 3DPW specifically designed to assess model robustness against significant occlusions. Besides, for 48- and 96-joint representations, we follow Pose2Mesh~\cite{choi2020pose2mesh} and apply HEM-based graph coarsening on the SMPL mesh to generate the corresponding skeletons. See more details in Appendix~\ref{sec:data_processing}.

\paragraph{Evaluation Metrics} For Human3.6M and 3DPW, we evaluate the model under two protocols: (1) MPJPE, mean Euclidean distance between predicted and GT joints after root joint alignment, and (2) PA-MPJPE (Procrustes-Aligned), which aligns predicted and GT joints via rigid transformation and then calculates MPJPE. For MPI-INF-3DHP, we employ MPJPE, percentage of correct keypoints (PCK) at 150 mm accuracy, and the area under the curve (AUC) as evaluation metrics to assess 3D lifting results. For 2D evaluation, we adopt root-aligned 2D Mean Error to focus on relative accuracy, as the root in the sparse skeleton is highly precise.

\subsection{Implementation Details}
\paragraph{Granularity-agnostic Pose Tokenizer (GPT)} 
GPT is built upon a hierarchical Vector Quantized Variational Autoencoder (VQ-VAE)~\cite{van2017neural} architecture with three encoder–decoder pairs for $\mathbf{x}_1$, $\mathbf{x}_2$, and $\mathbf{x}_3$ ($j_1 = 17$, $j_2 = 48$, and $j_3 = 96$). The Joint-to-Token expansion strategy with factor $6$ produces feature granularities $g_n = 6j_n$ for the corresponding latent features $\mathbf{Z}_n$. All levels share a unified codebook ($V\!=\!4096$, $D\!=\!128$) with quantization densities $\{\rho_k\}_{k=1}^{K} = [48, 102, 192, 288, 432, 576]$. Each encoder/decoder consists of $9$ MLP layers and $4$ MLP-Mixer~\cite{tolstikhin2021mlp} blocks. All interpolations in this paper are area-based unless otherwise specified. The GPT model is pretrained on Human3.6M for $100$ epochs using AdamW ($1e-3$) with cosine scheduling and $5$-epoch warmup.

\paragraph{Unified Skeletal Autoregressive Model (UniSkelar)} 
UniSkelar is a $4$-layer decoder-only Transformer ($D\!=\!256$, $16$ heads) that autoregressively models multi-density latent features $z_k$. RoPE is used for positional encoding. It is trained for 100 epochs with the same optimization settings as GPT. A block-wise causal attention mask enforces next-scale dependencies across granularities, and KV-caching accelerates inference. In granularity-agnostic experiments across unseen test poses ($\mathbf{x}_{1}^{\dagger}$, $\mathbf{x}_{2}^{\dagger}$, and $\mathbf{x}_{3}^{\dagger}$ with 192, 384, and 768 joints, respectively), only the lightweight decoders for these novel granularities are retrained as topology-specific projection heads, while the pretrained GPT encoder, shared codebook, and the UniSkelar model remain frozen.

\paragraph{3D Lifting}
Our densify-then-lift pipeline follows standard HPE evaluation protocols. We modify existing lifting models to take 161 joints (17+48+96) as input instead of the original 17 joints. As illustrated in Sec.~\ref{sec:lifting_plug_in_strategy}, 
joint-wise concatenation is used to retrain MixSTE on Human3.6M. Cross-attention fusion is used for GFPose and D3DP, which are fine-tuned from pretrained weights on Human3.6M (SH) and MPI-INF-3DHP (GT), respectively.

\subsection{SOTA in 2D Pose Densification}
\begin{figure}[t]
  \centering
  \includegraphics[width=\linewidth]{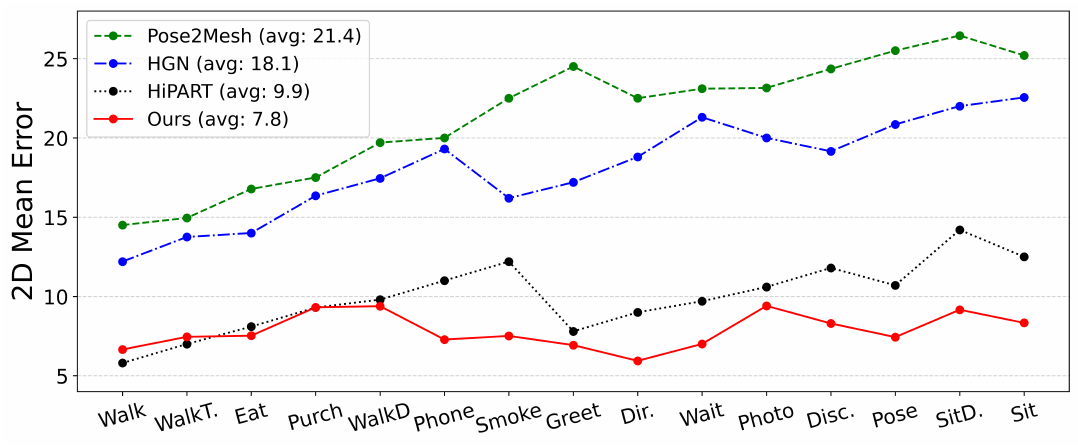}
  \caption{Action-wise densification results of four methods on Human3.6M with ground truth 17-joint 2D sparse inputs.}
  \Description{Action-wise densification results of four methods on Human3.6M with ground truth 17-joint 2D sparse inputs.}
  \label{fig:compare_2d}
\end{figure}

As shown in Figure~\ref{fig:compare_2d}, Pose2Mesh~\cite{choi2020pose2mesh} and HGN~\cite{li2021hierarchical} are not tailored for 2D densification and yield suboptimal results, highlighting the specificity required in this domain. VARPose achieves 7.8 px, surpassing the task-matched HiPART~\cite{zheng2025hipart} by 2.1 px (7.8 vs. 9.9 px), with notable gains on challenging actions such as \textit{Smoke} and \textit{SitDown}. These improvements demonstrate the effectiveness of our architecture, where GPT provides a unified codebook and UniSkelar leverages it for robust representation learning in pose densification. More densification results under CPN inputs and on 3DPW are included in Appendix~\ref{subsec:more_2d_densification_results}.

\subsection{Tangible Enhancement in 3D HPE}
\paragraph{Results on Human3.6M}
As shown in Table~\ref{tab:human3.6m_comparison}, under SH-detected 2D poses, VARPose built upon GFPose achieves 35.0 ($\Delta$0.6)~mm; under ground-truth 2D poses, VARPose combined with MixSTE ($f\!=\!81$) achieves 25.0 ($\Delta$0.9)~mm. These consistent improvements validate that dense joints provide fine-grained structural information that effectively benefits 3D pose lifting. Additional CPN-based results are reported in Appendix~\ref{subsec:more_details_about_3D_experiments}.

\paragraph{Results on MPI-INF-3DHP}
To evaluate generalization, we apply VARPose pretrained on Human3.6M directly to MPI-INF-3DHP for 2D pose densification, then refine D3DP ($f\!=\!243$) with the dense inputs. As shown in Table~\ref{tab:3dhp_results}, the consistent improvements across datasets confirm that the dense representations learned by VARPose generalize well.
\begin{table}[t]
  \centering
  \caption{Results on MPI-INF-3DHP under three evaluation metrics using ground-truth 2D keypoints as inputs.}
  \label{tab:3dhp_results}
  \resizebox{\columnwidth}{!}{
  \begin{tabular}{c|cccc}
    \hline
    Method &  \#Frames & PCK$\uparrow$ & AUC$\uparrow$ & MPJPE$\downarrow$ \\
    \hline
    MHFormer~\cite{li2022mhformer} &  9 & 93.8 & 63.3 & 58.0 \\
    MixSTE~\cite{zhang2022mixste} &  27 & 94.4 & 66.5 & 54.9 \\
    MixSTE~\cite{zhang2022mixste} &  243 & 96.9 & 75.8 & 35.4 \\
    D3DP~\cite{shan2023diffusion} $(H\!=\!1,K\!=\!1)$& 243& \underline{97.7}& \underline{77.8}& \underline{30.2}\\
    \hline
    D3DP$(H\!=\!1,K\!=\!1)$+VARPose(Ours) & 243 & \textbf{97.8}\textcolor{gray}{$\Delta$0.1} & \textbf{77.9}\textcolor{gray}{$\Delta$0.1} & \textbf{29.9}\textcolor{gray}{$\Delta$0.3} \\ 
    \hline
  \end{tabular}
  }
\end{table}
\begin{figure}[t]
    \centering
    \includegraphics[width=\linewidth]{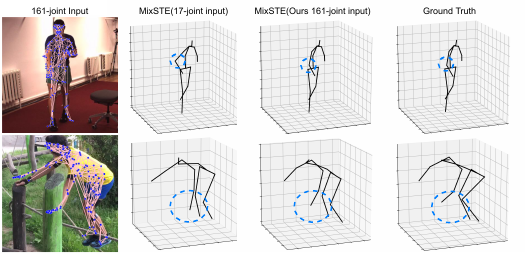}
    \caption{Qualitative results compared with MixSTE on Human3.6M (top) and 3DPW (bottom).}
    \Description{Qualitative results compared with MixSTE on Human3.6M (top) and 3DPW (bottom).}
    \label{fig:lifting_visualization}
\end{figure}
\paragraph{Qualitative Results} Figure~\ref{fig:lifting_visualization} (top) illustrates the 3D lifting results of MixSTE $(f\!=\!81)$ and our VARPose-enhanced MixSTE $(f\!=\!81)$ using Human3.6M ground truth inputs. The visualization demonstrates that by incorporating 2D dense poses, our framework effectively rectifies reconstruction artifacts in limb regions, even under occlusion conditions.

\subsection{Insights of VARPose}
\begin{figure}[t]
\centering
\includegraphics[width=\linewidth]{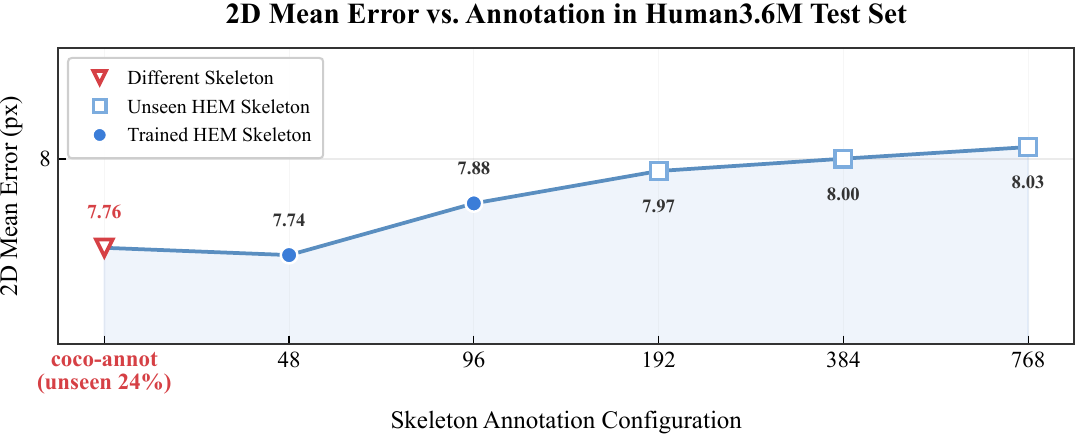}
\caption{Scalability and flexibility of the decoder.} 
\Description{Scalability and flexibility of the decoder.}
\label{fig:scale_trend}
\end{figure}

\paragraph{Effectiveness of Unified Tokenization and Next-scale Prediction}
Table~\ref{tab:varpose_insights} isolates the contributions of unified tokenization and next-scale prediction. First, our Unified Codebook significantly outperforms the Pose-Specific Codebook (PSC) baseline as detailed in Appendix~\ref{sec:more_insights_of_varpose}, reducing reconstruction and densification errors from 2.21\,px and 14.93\,px to 0.36\,px and 9.61\,px, respectively. Second, the next-scale paradigm further lowers the densification error from 9.61\,px to 7.81\,px compared to the next-token strategy. Collectively, the unified codebook captures shared inter-granularity correlations while the next-scale paradigm models hierarchical dependencies, jointly yielding superior pose representations.
\begin{table}[t]
\centering
\caption{Comparison of pose tokenization and autoregressive prediction on Human3.6M. Reconstruction and densification errors are measured in pixels.}
\label{tab:varpose_insights}
\resizebox{0.85\columnwidth}{!}{
\begin{tabular}{llcc}
\toprule
Tokenizer & AR Strategy & Reconstruction & Densification \\
\midrule
PSC & Next-token & 2.21 & 14.93 \\
GPT & Next-token & \textbf{0.36} & 9.61 \\
\graycell{GPT} & \graycell{Next-scale} &
\graycell{\textbf{0.36}} &
\graycell{\textbf{7.81}} \\
\bottomrule
\end{tabular}
}
\end{table}
\paragraph{Generating Novel, Unseen Poses}
VARPose generates comparable poses under unseen joint configurations as shown in Figure~\ref{fig:scale_trend}. Notably, the COCO skeleton, semantically distinct with 24\% unseen joints, achieves 7.76\,px, matching seen-granularity performance and confirming that the learned representation captures annotation-agnostic kinematic structure. The scaling trend is equally encouraging, with error increasing only marginally from 7.74 to 7.88\,px as joints scale from 48 to 96, and then stabilizing at finer granularities, reaching 7.97\,px, 8.00\,px, and 8.03\,px at 192, 384, and 768 joints, respectively. This plateau demonstrates strong scaling potential for finer pose representations. 

\medskip
Additional information, such as implementation details and qualitative analysis of pose-specific versus unified codebooks, feature-space analysis of semantic alignment, and few-shot adaptation capability, is provided in Appendix~\ref{sec:more_insights_of_varpose}.

\subsection{Ablation Studies}
\paragraph{2D Dense Pose Effect on 3D Downstream Tasks}
We ablate VARPose across 3D HPE and HMR under multiple input settings. As shown in Table~\ref{tab:2d_dense_pose_effect_3d_lifting} and Table~\ref{tab:3d_hmr}, VARPose consistently outperforms the 17-GT baseline (HPE: 25.0 vs.\ 25.9\,mm under MixSTE; HMR: MPJPE-24 67.4 vs.\ 69.5\,mm) and is much closer to the full 161-joint ground truth.
Moreover, as reported in Table~\ref{tab:3d_hpe_across_different_joints}, increasing the number of input joints from 17 to 65 (17+48) and further to 161 (17+48+96) yields progressive gains in both MPJPE (35.6\,$\rightarrow$\,35.3\,$\rightarrow$\,35.0\,mm) and PA-MPJPE (30.5\,$\rightarrow$\,30.0\,$\rightarrow$\,29.9\,mm), confirming that richer dense pose input continuously benefits 3D downstream tasks.
However, the gains are not due to joint count alone: under the vanilla transformer setting in Appendix Table~\ref{tab:vanilla_lifting}, VARPose outperforms bone interpolation with the same input dimensionality (MPJPE: 38.9 vs.\ 41.9\,mm; PA-MPJPE: 30.6 vs.\ 32.3\,mm), confirming the importance of anatomically coherent structure.
Lifting models with VARPose dense input attend to fine-grained anatomical regions, indicating that the learned dense pose encodes a structured body prior that transfers across downstream tasks.
\begin{table}[t]
\centering
\caption{Effect of 2D dense pose on 3D HPE using MPJPE.}
\label{tab:2d_dense_pose_effect_3d_lifting}
\setlength{\tabcolsep}{1pt}
\begin{tabular*}{0.9\columnwidth}{@{\extracolsep{\fill}}lcc}
\toprule
\multirow{2}{*}{Input Setting} & MixSTE$(f\!=\!81)$  & GFPose $(S\,=\,200)$ \\
\cmidrule(r){2-3}
 & Concatenation & Cross Attention \\
\midrule
17-GT              & 25.9  & 18.9 \\
161-VARPose        & \underline{25.0}  & \underline{18.5} \\
161-GT             & \textbf{16.8}  & \textbf{18.4} \\
\bottomrule
\end{tabular*}
\end{table}
\begin{table}[t]
\centering
\caption{Effect of 2D dense pose on 3D HMR with MixSTE as a lifting backbone.}
\label{tab:3d_hmr}
\setlength{\tabcolsep}{1pt}
\begin{tabular*}{0.9\columnwidth}{@{\extracolsep{\fill}}lcccc}
\toprule
Input Setting & MPJPE-24 & PA-MPJPE-24 & MPVPE & PA-MPVPE \\
\midrule
17-GT              & 69.5 & 31.2 & 78.3 & 39.8 \\
161-VARPose & 67.4 & 29.9 & 75.8 & 38.8 \\
161-GT             & \underline{61.4} & \underline{26.0} & \underline{67.8} & \underline{34.2} \\
17-GT-3D           & \textbf{58.3} & \textbf{23.1} & \textbf{65.1} & \textbf{32.2} \\
\bottomrule
\end{tabular*}
\end{table}

\begin{table}[t]
\centering
\caption{3D HPE performance across different joints input.}
\label{tab:3d_hpe_across_different_joints}
\renewcommand{\arraystretch}{1.15}
\begin{tabular*}{0.9\columnwidth}{@{} l @{\extracolsep{\fill}} c @{\hspace{1.5em}} c @{}}
  \toprule
  GFPose ($S\!=\!200$; Human3.6M(SH)) & MPJPE & PA-MPJPE \\
  \midrule
  w/o VARPose (17)                    & 35.6           & 30.5              \\
  w/ VARPose (17+48)                  & \underline{35.3\,($\Delta$0.3)} & \underline{30.0\,($\Delta$0.5)} \\
  w/ VARPose (17+48+96)               & \textbf{35.0\,($\Delta$0.6)}   & \textbf{29.9\,($\Delta$0.6)}   \\
  \bottomrule
\end{tabular*}
\end{table}

\paragraph{In-the-Wild Robustness}
We further evaluate VARPose on the challenging benchmark 3DPW with CPN-detected 2D inputs, including the 3DPW-Occ subset that reflects real-world occlusions. The results show consistently improved performance on both datasets, with especially notable gains on 3DPW-Occ. As shown in Table~\ref{tab:3dpw_results}, injecting dense poses reduces error under occlusion, confirming the practical use of 2D pose densification. The qualitative result is shown in Figure~\ref{fig:lifting_visualization} (bottom). 

\medskip
Additional ablations on model components, hyperparameters, and token expansion, together with stress tests under diverse input degradations, are provided in Appendix~\ref{sec:more_ablation}.

\subsection{Computational Efficiency}
\label{sec:computational_efficiency}
As summarized in Table~\ref{tab:computational_efficiency}, the full pipeline \emph{\hl{approaches real-time}} on a single 48\,GB RTX\,4090. Under pipeline scheduling, the two stages execute concurrently, so \textbf{the system latency} equals the bottleneck: $\max(\text{lat}_{\text{VARPose}},\; \text{lat}_{\text{down}}/f) = $ \hlc{bgyellow}{21.6\,ms/frame}. Specifically, the 21.6\,ms breaks down into 3.4\,ms input processing and 6 autoregressive steps of 2.3, 2.6, 3.1, 3.2, 3.2, 3.8\,ms. This is achieved despite higher MACs (8.9 vs.\ 0.3\,G) and memory (277.0 vs.\ 38.4\,MB) than HiPART, because parallel next-scale prediction replaces sequential token generation, yielding \underline{3.4$\times$ faster} inference and \underline{2.1\,px lower} error. For downstream lifting, the plug-in strategy determines the overhead. Concatenation on MixSTE incurs higher MACs (439.0 vs.\ 46.4\,G) and latency (39.3 vs.\ 5.3\,ms) but brings 0.9\,mm gains with faster convergence (44 vs.\ 81 epochs), while cross-attention on D3DP adds negligible cost (+0.08\%\,MACs, +3.2\,ms) and still improves by 0.3\,mm in only 36 fine-tuning epochs. Both plug-in strategies add only about 270\,MB of memory. Overall, VARPose is not only effective but also practically efficient and deployable.

\begin{table}[t]
\centering
\caption{Results on 3DPW and 3DPW-Occ.}
\label{tab:3dpw_results}
\resizebox{\linewidth}{!}{
\begin{tabular}{lcccc}
\toprule
\multirow{2}{*}{Method} & \multicolumn{2}{c}{3DPW-Occ} & \multicolumn{2}{c}{3DPW}  \\
\cmidrule(r){2-3} \cmidrule(r){4-5} & MPJPE & PA-MPJPE & MPJPE & PA-MPJPE \\
\midrule
PoseFormer \cite{zheng20213d} & 132.8 & 80.5 & 118.2 & 73.1 \\
MixSTE \cite{zhang2022mixste} & 121.5 & 75.5 & 118.9 & 73.7  \\
\midrule
MixSTE+VARPose(Ours) & \textbf{118.3$\Delta$3.2}& \textbf{74.0$\Delta$1.5}  & \textbf{118.4$\Delta$0.5} & \textbf{72.7$\Delta$1.0} \\ \bottomrule
\end{tabular}
}
\end{table}
\begin{table}[t]
  \centering
  \caption{Computational efficiency. Lat.\&Mem at bs=1 on RTX 4090. $^*$: downstream only. $\dagger$: reproduced.}
  \label{tab:computational_efficiency}
  \setlength{\tabcolsep}{2pt}
  \resizebox{\columnwidth}{!}{
  \begin{tabular}{lcccccc}
    \toprule
    Method & $f$ & MACs(G) & Params(M) & \multicolumn{1}{c}{Mem(MB)} & \multicolumn{1}{c}{Lat.(ms)} & \multicolumn{1}{c}{Epoch} \\
    \midrule
    \multicolumn{7}{l}{\textit{2D Densification (per frame)}} \\
    HiPART$^{\dagger}$ & 1 & 0.33 & 2.4 & 38.4 & 72.7 & N/A \\
    VARPose & 1 & 8.9 & 22.7 & 277.0 & 21.6 & 61  \\
    \midrule
    \multicolumn{7}{l}{\textit{3D Lifting Downstream (per sample)$^*$}} \\
    MixSTE & 81 & 46.35 & 33.7 & 173.8 & 5.3 & 81  \\
    \quad+VARPose & 81 & 439.02 & 33.7 & 443.4 & 39.3 & 44  \\
    D3DP & 243 & 139.05 & 34.7 & 272.7 & 13.7 & 400  \\
    \quad+VARPose & 243 & 139.16 & 34.7 & 534.3 & 16.9 & 36 \\
    \bottomrule
  \end{tabular}
  }
\end{table}
\section{Conclusion}
This paper introduces VARPose, a framework that advances 3D human pose estimation by transposing the successful ``next-scale prediction'' paradigm from image generation to this domain for the first time. VARPose not only achieves substantial accuracy gains on trained (seen) pose granularities but also generates new, unseen granularities, demonstrating a new problem setting in this field. Crucially, we demonstrate that these generated dense poses serve as powerful intermediate representations that enrich structural context for downstream tasks. Through our two proposed plug-in strategies, VARPose effectively rectifies structural artifacts in existing 3D lifting backbones, particularly under severe occlusion.

We believe this unified generative framework provides a scalable foundation for diverse downstream applications such as fine-grained action analysis and human mesh recovery, and will inspire further exploration of skeletal foundation models.

\begin{acks}
This work was supported by the National Natural Science Foundation of China (Grant No. 62206123) and the Key Science and Technology Program of Lhasa Municipality (Grant No. LSKJ202612).
\end{acks}

\bibliographystyle{ACM-Reference-Format}
\balance
\bibliography{varpose}

\clearpage
\newpage
\nobalance

\appendix

\section{Overview}
\label{sec:overview}
This supplementary material provides additional details, experiments, and analyses to complement the main paper. The sections are organized as follows:
\begin{itemize}[noitemsep, nosep, left=0pt]
    \item \textbf{\cref{sec:data_processing}: Data Processing.} Construction of our pipeline for a hierarchical pose dataset via mesh coarsening and motion capture indexing.
    
    \item \textbf{\cref{sec:more_insights_of_varpose}: More Insights of VARPose.} Implementation details and qualitative analysis of pose-specific versus unified codebooks, feature-space analysis of semantic alignment, and few-shot adaptation capability.

    \item \textbf{\cref{sec:additional_experiment}: Additional Experiment Results.} Extended evaluations on CPN-based densification, cross-dataset generalization to 3DPW, 3D HPE lifting with various backbones, input-enrichment comparison with AugLift, and H36M-to-SMPL translation for HMR.

    \item \textbf{\cref{sec:more_ablation}: More Ablation Studies.} GPT module effectiveness, hyperparameter selections, joint-to-token expansion strategies (balanced vs. position-only), codebook sensitivity analysis (vocabulary size and embedding dimension), and RoPE necessity for skeletal topology, as well as robustness under Gaussian noise, detection errors, extreme poses, and motion blur.

    \item \textbf{\cref{sec:more_visualization}: More Visualization Analysis.} Qualitative 2D reconstructions across multiple granularities (17--768 joints) demonstrate graceful degradation and 3D downstream improvements in HPE and HMR tasks.

    \item \textbf{\cref{sec:limitation}: Limitations and Future Work.} Lack of temporal modeling, suboptimal plug-in efficiency, and fundamental scope boundaries regarding non-rigid body parts and multi-person scenarios.
\end{itemize}

\section{Data Processing}
\label{sec:data_processing}
\begin{figure}[t]
\centering
\includegraphics[width=\linewidth]{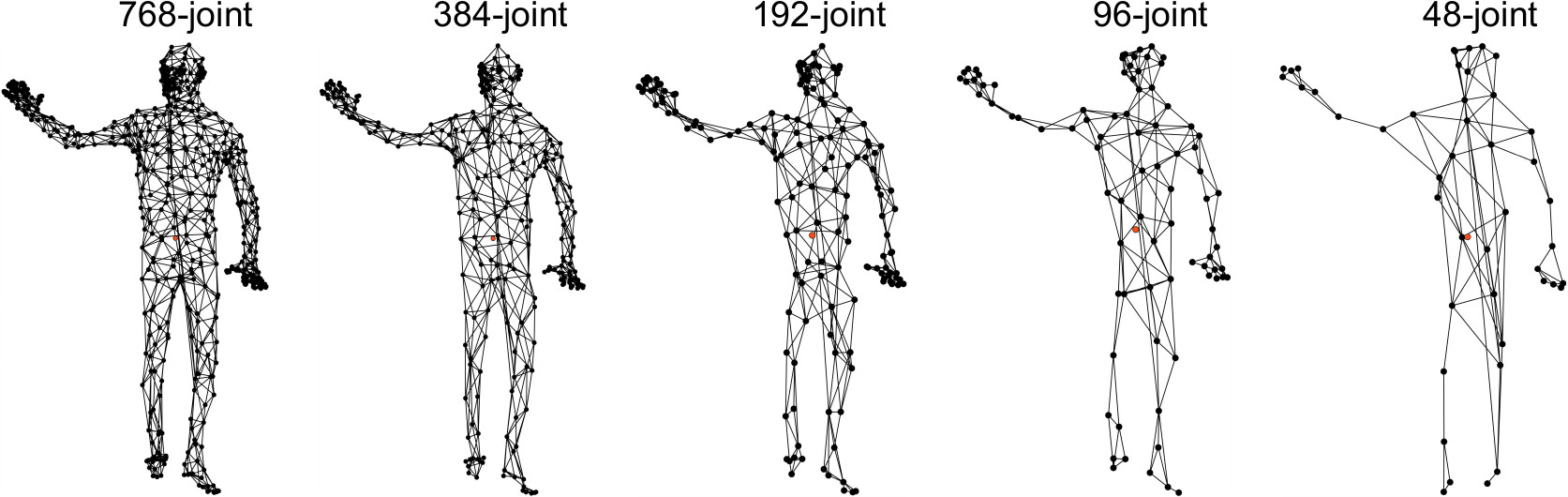}
\caption{Skeletons derived from mesh coarsening. The central red dot highlights the repositioned pseudo-nodes at the root joint.}
\Description{Skeletons derived from mesh coarsening. The central red dot highlights the repositioned pseudo-nodes at the root joint.}
\label{fig:768_384_192_96_48_skeletons}
\end{figure}

Given the absence of a public benchmark for hierarchical 2D human poses, our data generation pipeline consists of two complementary categories: mesh-based and motion capture-based representations. For the mesh-based category, we leverage the pseudo ground truth 3D human meshes from Pose2Mesh, based on the SMPL model with 6890 vertices. We apply the Heavy Edge Matching (HEM) \cite{defferrard2016convolutional} algorithm for iterative graph coarsening, generating five representations with 768, 384, 192, 96, and 48 nodes, respectively, as illustrated in Figure \ref{fig:768_384_192_96_48_skeletons}. A key characteristic of this coarsening process is the introduction of pseudo-nodes: since the HEM algorithm constructs a binary tree for hierarchical pooling, each parent node must have exactly two children. When a single node cannot be paired, a pseudo-node is synthetically created as its sibling, meaning our coarsened node sets contain a mixture of original mesh vertices and synthetic pseudo-nodes lacking physical locations.

\begin{figure*}[t]
    \centering
    \includegraphics[width=0.9\linewidth]{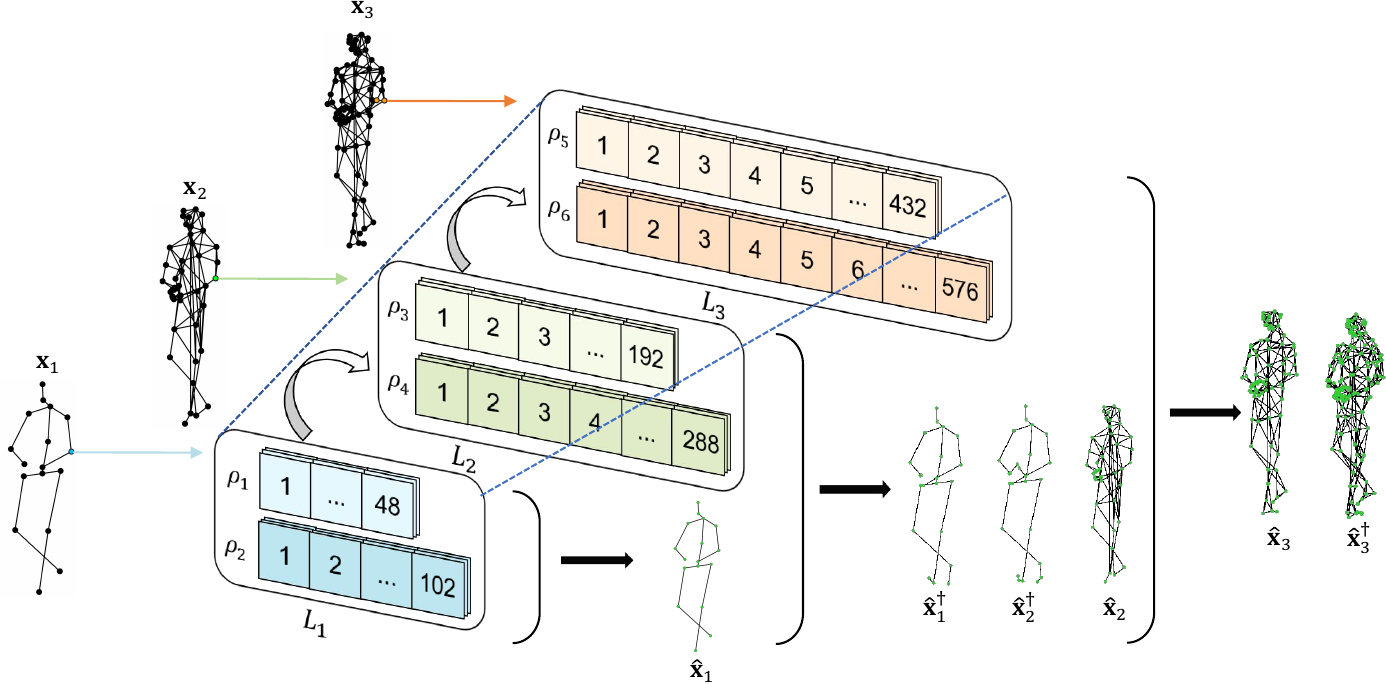}
    \caption{Granularity-agnostic Pose Tokenizer Procedure. It demonstrates that quantized features $z_k$ corresponding to intermediate quantization densities $\rho_k$ still explicitly encode meaningful and generalizable pose representations, enabling generation at flexible granularities, including novel poses $\mathbf{\hat{x}}^\dagger_{1,2,3}$.}
    \Description{Granularity-agnostic Pose Tokenizer Procedure. It demonstrates that quantized features $z_k$ corresponding to intermediate quantization densities $\rho_k$ still explicitly encode meaningful and generalizable pose representations, enabling generation at flexible granularities, including novel poses $\mathbf{\hat{x}}^\dagger_{1,2,3}$.}
    \label{fig:GPT}
\end{figure*}

For the motion capture-based category, we directly index subsets of the original 32 joints from the Human3.6M dataset. Analysis reveals multiple perfectly co-located keypoints, reducing the set to 25 unique joints $\mathbf{\hat{x}}^{\dagger}_2$. Removing four extremity keypoints ($[4, 9, 21, 29]$) yields a 21-joint skeleton $\mathbf{\hat{x}}^{\dagger}_1$. The standard 17-joint skeleton from VideoPose3D serves as the sparsest level. As shown in Figure \ref{fig:GPT}, these three granularities (25, 21, and 17 joints) are grounded in high-quality motion capture data. The GT and CPN-detected poses on Human3.6M annotations are sourced from VideoPose3D, and the SH-detected data are obtained from GFPose. Additionally, we generate pseudo ground truth on COCO annotations using CPN to enable training and evaluation of the generalizable decoder. 

For the training and evaluation of our model, we select a specific three-level configuration from our generated dataset: the 96-joint and 48-joint skeletons from the mesh coarsening process, and the 17-joint GT skeleton. \textbf{A critical challenge in using the coarsened graphs is the handling of the aforementioned pseudo-nodes.} To address this, we adopt an effective strategy during both training and inference: we relocate all pseudo-nodes present to the exact 3D coordinate of the root joint (pelvis) of the corresponding 17-joint ground-truth pose. This ensures that the pseudo-nodes do not contribute erroneously to the geometric loss function or visual output, while preserving the integrity of the graph topology.

\section{More Insights of VARPose}
\label{sec:more_insights_of_varpose}
\paragraph{Pose-Specific vs.\ Unified Codebooks}
To validate our Granularity-agnostic Pose Tokenizer (GPT), we establish a Pose-Specific Codebook (PSC) baseline based on HiPART. PSC employs a dual-codebook VQ-VAE to disentangle poses into sparse ($C_s$) and dense ($C_d$) spaces. The sparse latent $z_s$ is quantized by $C_s$, and the result guides dense quantization. To align the codebooks, we implement a graph-aware mapping strategy. Specifically, after discarding the root joint, each sparse joint $s_i$ is mapped to its nearest dense keypoint $d_{\text{anchor}}$ via Euclidean distance. We then restrict the search to local neighbors using a predefined adjacency matrix to form triplet correspondences for the Info-NCE loss.

\begin{figure*}[t]
    \centering
    \includegraphics[width=0.9\linewidth]{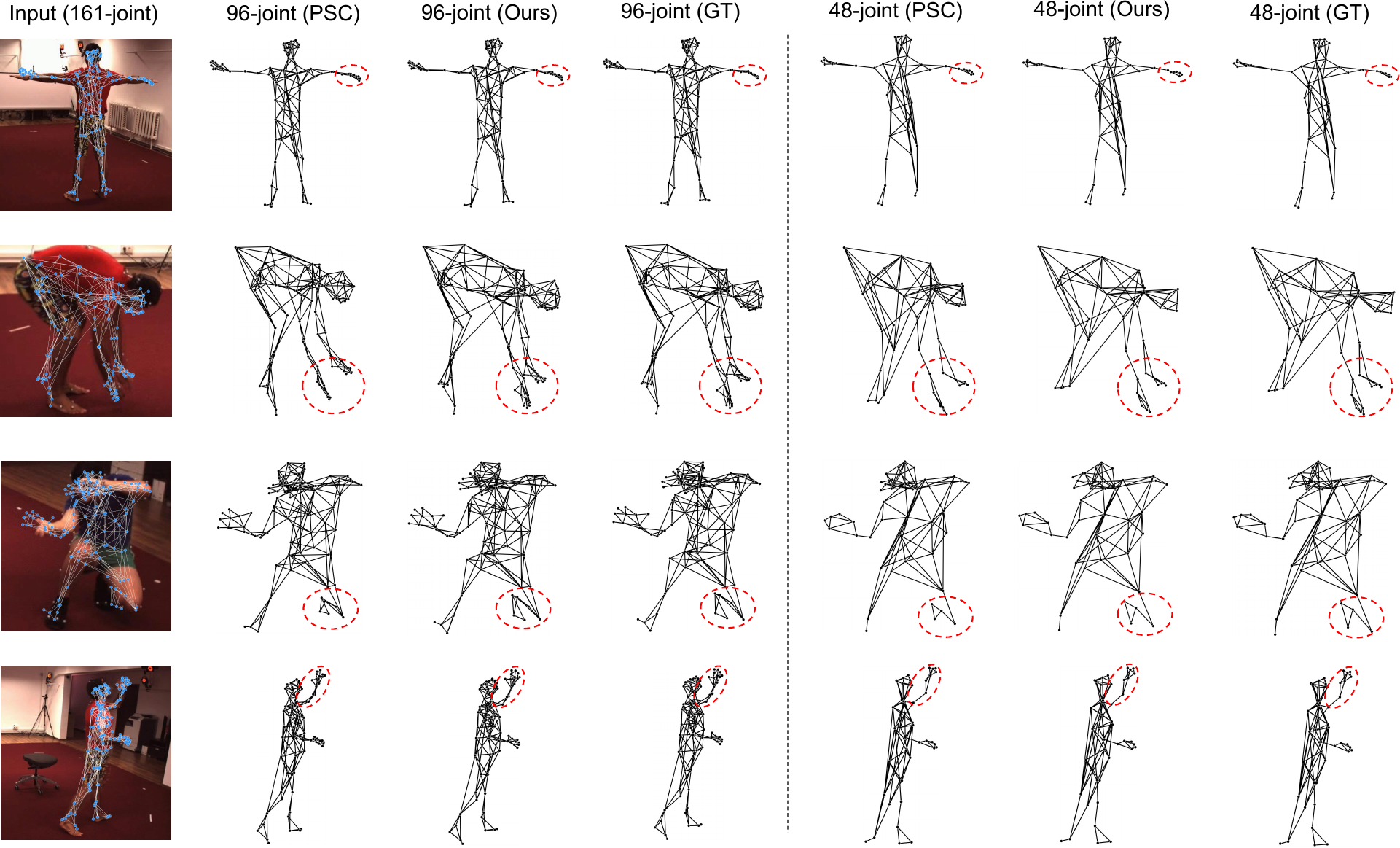}
    \caption{Qualitative comparison of 2D pose reconstructions on the Human3.6M dataset. For both 96-joint and 48-joint granularities, we compare the results from the Pose-Specific Codebook (PSC) baseline and our Unified Codebook against the Ground Truth (GT). }
    \Description{Qualitative comparison of 2D pose reconstructions on the Human3.6M dataset. For both 96-joint and 48-joint granularities, we compare the results from the Pose-Specific Codebook (PSC) baseline and our Unified Codebook against the Ground Truth (GT). }
    \label{fig:PSC_comparison}
\end{figure*}

Compared to PSC's isolated discrete spaces, our GPT employs a unified codebook $\mathcal{C}$ that learns a continuous anatomical manifold across diverse granularities, as illustrated in Figure~\ref{fig:GPT}. Qualitative comparisons in Figure~\ref{fig:PSC_comparison} further confirm that GPT achieves greater robustness on challenging samples, demonstrating that a shared pose dictionary yields more robust representations than domain-specific counterparts.

\begin{figure}[t]
  \centering
  \includegraphics[width=0.8\linewidth]{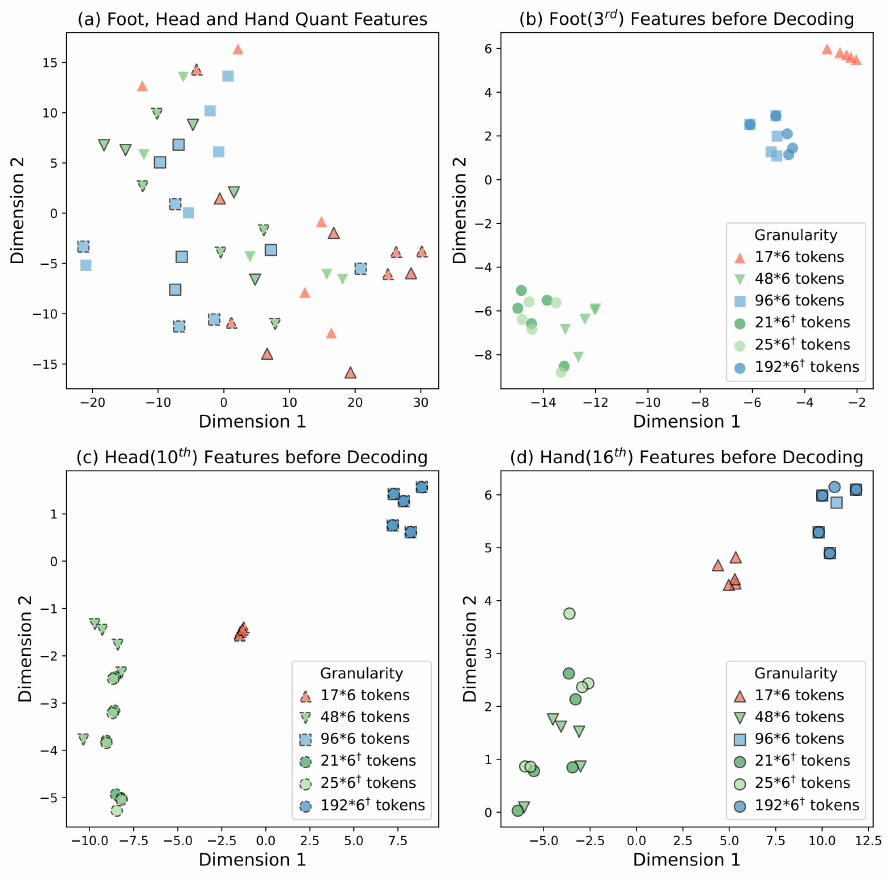}
  \caption{Feature visualization across different pose granularities. (a) The unified codebook produces scattered quantized features, indicating granularity independence. (b--d) Trained and unseen poses cluster clearly before decoding, highlighting our generalization.
  }
  \Description{Feature visualization across different pose granularities. (a) The unified codebook produces scattered quantized features, indicating granularity independence. (b--d) Trained and unseen poses cluster clearly before decoding, highlighting our generalization.}
  \label{fig:feature_space_analysis}
\end{figure}

\paragraph{Granularity-Agnostic Codebook and Decoder Semantic Alignment}
Feature space analysis directly explains this generalization. As shown in Figure~\ref{fig:feature_space_analysis} (a), codebook vectors across quantization densities $\rho_k$ show no granularity-dependent clustering, indicating the codebook learns granularity-invariant pose primitives. In contrast, pre-decoding features cluster by target fusion resolution $\rho_k$ as shown in Figure~\ref{fig:feature_space_analysis} (b--d): e.g., sparse $g_1^{\dagger}=21$ (targeting $\rho_{K_1}$) aggregates with dense $g_2=48$, which supplies codebook vectors $z_{K_1}$. This reveals that the decoder unifies multi-granularity codebook entries into a shared semantic space, enabling reconstruction of never-encoded poses, which is the core mechanism behind cross-granularity generalization.

\begin{wraptable}[7]{r}{0.5\columnwidth}
  \captionsetup{aboveskip=0pt, belowskip=2pt}
  \centering
  \caption{Few-shot adaptation.}
  \label{tab:less_training_samples}
  \setlength{\tabcolsep}{3pt}
  \resizebox{\linewidth}{!}{
  \begin{tabular}{lc}
    \toprule
    Size Ratio & 2D Mean Error (px) $\downarrow$  \\
    \midrule
    30\%    & 8.03 ($\Delta$0.27) \\
    50\%    & 7.92 ($\Delta$0.16) \\
    \graycell{\textbf{100\%}}& \graycell{\textbf{7.76 ($\Delta$0.00)}}  \\
    \bottomrule
  \end{tabular}
  }
\end{wraptable}
\paragraph{Few-shot Adaptation} Under the representation-projection paradigm, the codebook learns topology-agnostic pose primitives while a lightweight decoder handles topology-specific coordinate mapping. This decoupling is empirically validated by adapting only the 450.55K-param decoder to the unseen COCO 17-joint skeleton using merely 30\% training data, which adds just 0.27~px error as shown in Table~\ref{tab:less_training_samples}, demonstrating that the shared representation generalizes without relearning.

\section{Additional Experiment Results}
\label{sec:additional_experiment}
\subsection{More 2D densification results}
\label{subsec:more_2d_densification_results}
\paragraph{Human3.6M CPN Densification}
In our primary experiments, VARPose is trained using ground-truth 2D poses for all hierarchical levels (17, 48, and 96 joints). However, in practical applications, 17-joint sparse inputs are typically obtained from 2D detectors such as CPN. To bridge this domain gap and evaluate the model's denoising and rectification capabilities, we conduct a mixed-source experiment on Human3.6M. We maintain the 48-joint and 96-joint data as ground truth to serve as high-fidelity densification targets while replacing the 17-joint GT poses with CPN-detected poses. By keeping all other hyperparameters and the hierarchical architecture constant, we retrain both the GPT tokenizer and the UniSkelar AR model. As a result, VARPose achieves a 2D Mean Error of \textbf{11.16}~px on the Human3.6M test set. While this error is slightly higher than that of the pure GT-trained version (\textbf{7.81}~px), it demonstrates superior robustness when generalizing to other datasets that rely on CPN detections.
\begin{figure}[t]
\centering
\includegraphics[width=0.9\linewidth]{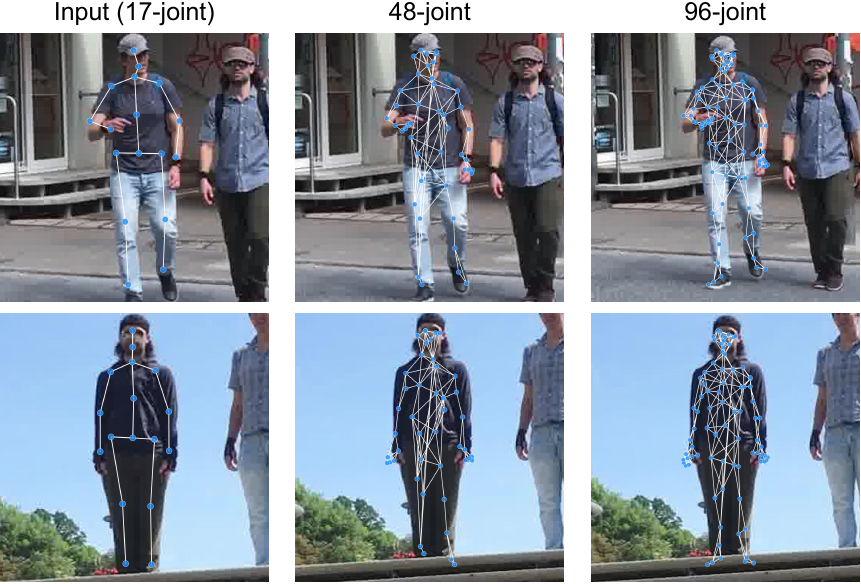}
\caption{Illustration of densification on 3DPW GT test set.} 
\Description{Illustration of densification on 3DPW GT test set.}
\label{fig:3dpw_densification_demo}
\end{figure}
\paragraph{3DPW Densification}
We apply the model trained on Human3.6M directly to the 3DPW ground truth test set without fine-tuning, achieving a 2D Mean Error of \textbf{23.96}\,px. As a real-world benchmark, 3DPW inherently features natural occlusions such as self-occlusions, object interactions, and out-of-view joints. As shown in Figure~\ref{fig:3dpw_densification_demo}, our unified codebook encodes cross-scale structural dependencies, enabling the ``next-scale prediction'' paradigm to propagate reliable cues from visible to occluded regions. This validates the transferability of our approach to unconstrained, in-the-wild environments.

\begin{table}[t]
    \centering
    \caption{Ablation study on vanilla transformer for lifting on Human3.6M GT test set.}
    \label{tab:vanilla_lifting}
    \resizebox{0.7\linewidth}{!}{
        \begin{tabular}{lcc}
            \toprule
            {Input Setting} & MPJPE & PA-MPJPE \\
            \midrule
            Bone Interpolation & 41.9 & 32.3 \\
            17-GT & 44.1 & 33.6 \\
            \graycell{161-VARPose} & \graycell{\underline{38.9}} & \graycell{\underline{30.6}} \\
            161-GT  & \textbf{17.9} & \textbf{14.2} \\
            \bottomrule
        \end{tabular}
    }
\end{table}
\subsection{More Details about 3D Experiments}
\label{subsec:more_details_about_3D_experiments}
\paragraph{Human3.6M CPN Results}
To further evaluate whether VARPose generalizes across different 2D detectors, we conduct experiments on the CPN detector, which is widely adopted in 3D human pose estimation. Specifically, we fine-tune FinePose~\cite{xu2024finepose} ($f\!=\!243$; $H\!=\!20$; $K\!=\!10$) with dense joints generated by VARPose and evaluate using the joint-aggregation strategy~\cite{shan2023diffusion}. As reported in Table~\ref{tab:human3.6m_cpn_comparison}, our approach achieves an average MPJPE of \textbf{39.8}\,mm, outperforming the original FinePose (40.2\,mm) by \textbf{0.4}\,mm and ranking first in 9 out of 15 individual actions. The most pronounced improvements are observed in \emph{SitD.}~($\Delta$2.1), \emph{Greet}~($\Delta$1.0), \emph{Disc.}~($\Delta$0.9), and both \emph{Dir.} and
\emph{Sit}~($\Delta$0.8). These results confirm that VARPose consistently enhances downstream 3D pose estimation performance regardless of the underlying 2D detector.

\begin{table*}[t]
  \centering
  \caption{Quantitative comparison with state-of-the-art methods on Human3.6M (CPN) using MPJPE. The best and second-best results are highlighted in \textbf{bold} and \underline{underlined}.}
  \label{tab:human3.6m_cpn_comparison}
  \resizebox{\textwidth}{!}{
  \begin{tabular}{l|ccccccccccccccc|c}
    \toprule
    Method & Dir. & Disc. & Eat & Greet & Phone & Photo & Pose & Purch. & Sit & SitD. & Smoke & Wait & WalkD. & Walk & WalkT. & Avg. \\
    \midrule
    VideoPose \cite{pavllo20193d}$(f\!=\!243)$ & 45.1 & 47.4 & 42.0 & 46.0 & 49.1 & 56.7 & 44.5 & 44.4 & 57.2 & 66.1 & 47.5 & 44.8 & 49.2 & 32.6 & 34.0 & 47.1 \\
    GraphSH \cite{xu2021graph} & 45.2 & 49.9 & 47.5 & 50.9 & 54.9 & 66.1 & 48.5 & 46.3 & 59.7 & 71.5 & 51.4 & 48.6 & 53.9 & 39.9 & 44.1 & 51.9 \\
    PoseFormer \cite{zheng20213d}$(f\!=\!81)$ & 41.5 & 44.8 & 39.8 & 42.5 & 46.5 & 51.6 & 42.1 & 42.0 & 53.3 & 60.7 & 45.5 & 43.3 & 46.1 & 31.8 & 32.2 & 44.3 \\
    MHFormer \cite{li2022mhformer}$(f\!=\!351)$ & \underline{39.2} & 43.1 & 40.1 & 40.9 & 44.9 & 51.2 & 40.6 & \underline{41.3} & 53.5 & 60.3 & 43.7 & 41.1 & 43.8 & 29.8 & 30.6 & 43.0 \\
    MixSTE \cite{zhang2022mixste}$(f\!=\!81)$ & 39.8 & 43.0 & 38.6 & 40.1 & 43.4 & 50.6 & 40.6 & 41.4 & 52.2 & 56.7 & 43.8 & 40.8 & 43.9 & 29.4 & 30.3 & 42.4 \\
    POT \cite{li2023pose} & 47.9 & 50.0 & 47.1 & 51.3 & 51.2 & 59.5 & 48.7 & 46.9 & 56.0 & 61.9 & 51.1 & 48.9 & 54.3 & 40.0 & 42.9 & 50.5 \\
    DiffPose \cite{gong2023diffpose} & 42.8 & 49.1 & 45.2 & 48.7 & 52.1 & 63.5 & 46.3 & 45.2 & 58.6 & 66.3 & 50.4 & 47.6 & 52.0 & 37.6 & 40.2 & 49.7 \\
    Di$^2$Pose \cite{wang2024text} & 41.9 & 47.8 & 45.0 & 49.0 & 51.5 & 62.2 & 45.7 & 45.6 & 57.6 & 67.1 & 50.1 & 45.3 & 51.4 & 37.3 & 40.9 & 49.2 \\
    Lifting by Image \cite{zhou2024lifting} & 44.9 & 46.4 & 42.4 & 44.9 & 48.7 & \textbf{40.1} & 44.3 & 55.0 & 58.9 & \textbf{47.1} & 48.2 & 42.6 & \textbf{36.9} & 48.8 & 40.1 & 46.4 \\
    FinePose \cite{xu2024finepose}$(f\!=\!243)$ & 39.5 & \underline{40.9} & \underline{35.8} & \underline{38.5} & \underline{42.0} & \underline{45.7} & 38.5 & \textbf{38.1} & 51.6 & 55.0 & 41.8 & 40.2 & 40.0 & \underline{27.9} & \underline{28.0} & \underline{40.2} \\
    HiPART \cite{zheng2025hipart} & 42.8 & 42.7 & 38.1 & 41.3 & 42.7 & 46.3 & \textbf{37.2} & 44.2 & \underline{51.0} & \underline{51.4} & \textbf{40.9} & \textbf{38.3} & 40.0 & 39.9 & 33.7 & 42.0 \\
    \midrule
    VARPose(Ours)$(f\!=\!243)$ & \textbf{38.7}\textcolor{gray}{$\Delta$0.8} & \textbf{40.0}\textcolor{gray}{$\Delta$0.9} & \textbf{35.7}\textcolor{gray}{$\Delta$0.1} & \textbf{37.5}\textcolor{gray}{$\Delta$1.0} & \textbf{41.9}\textcolor{gray}{$\Delta$0.1} & 46.2\textcolor{gray}{$\Delta$-0.5} & \underline{38.1}\textcolor{gray}{$\Delta$0.4} & \textbf{38.1}\textcolor{gray}{$\Delta$0.0} & \textbf{50.8}\textcolor{gray}{$\Delta$0.8} & 52.9\textcolor{gray}{$\Delta$2.1} & \underline{41.6}\textcolor{gray}{$\Delta$0.2} & \underline{40.1}\textcolor{gray}{$\Delta$0.1} & \underline{39.3}\textcolor{gray}{$\Delta$0.7} & \textbf{27.8}\textcolor{gray}{$\Delta$0.1} & \textbf{27.7}\textcolor{gray}{$\Delta$0.3} & \textbf{39.8}\textcolor{gray}{$\Delta$0.4} \\ 
    \bottomrule
  \end{tabular}
  }
\end{table*}

\paragraph{Vanilla Transformer for Lifting}
We employ a lightweight vanilla Transformer (12 blocks, 6 heads, 96 embedding dimensions) as the feature extractor, followed by a linear regression head, trained with a learning rate of $1e-4$ and cosine positional embedding. We evaluate four input settings: \emph{Bone Interpolation} linearly interpolates along the Human3.6M skeleton; \emph{17-GT} uses the original 17 Human3.6M ground-truth joints; \emph{161-VARPose} combines 17 GT joints with 48 and 96 joints estimated by VARPose; and \emph{161-GT} provides all 161 ground-truth joints. As reported in Table~\ref{tab:vanilla_lifting}, 161-VARPose yields 38.9 MPJPE / 30.6 PA-MPJPE, outperforming both Bone Interpolation (41.9 / 32.3) and 17-GT (44.1 / 33.6). With 161-GT, the model achieves 17.9 / 14.2. These results suggest that the fundamental bottleneck in existing 3D HPE lies in sparse skeletons that underutilize topology, and that dense joints with reasonable topology rather than higher density alone can effectively alleviate it.

\begin{figure}[t]
\centering
\includegraphics[width=\linewidth]{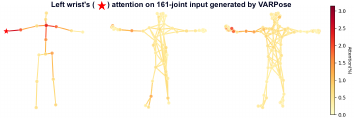}
\caption{Consistent enrichment of structural context.} 
\Description{Consistent enrichment of structural context.}
\label{fig:attention_heatmap}
\end{figure}
\paragraph{More Details about Lifting Visualization}
The attention heatmap presented in the main paper is derived from the proposed dual-channel fusion visualization method, applied to the vanilla transformer trained on the \emph{161-GT} setting. Specifically, Channel~A extracts the effective regression weight $W_{\text{eff}} \in \mathbb{R}^{17 \times 161}$ from the regression head, while Channel~B captures the average self-attention map $\bar{A} \in \mathbb{R}^{161 \times 161}$ across all 12 transformer layers on real test samples. The attention importance is then obtained by propagating $\bar{A}$ through $W_{\text{eff}}$ via a dot-product: $\text{attn\_imp}[k,j] = \sum_i \hat{W}_{\text{eff}}[k,i] \cdot \bar{A}[i,j]$, and fused with the regression importance through weighted summation with factor $\alpha$, yielding a joint-level importance score per output joint. An additional visualization result under the \emph{161-VARPose} setting is provided in Figure~\ref{fig:attention_heatmap}, which consistently demonstrates that the lifting model preferentially attends to denser joint regions, thereby aggregating fine-grained structural information for accurate 3D pose recovery.

\paragraph{H36M-to-SMPL Translator in 3D HMR}
SMPL-IKS requires an SMPL-24 skeleton as input, whereas most off-the-shelf 3D HPE models output predictions in the Human3.6M 17-joint format. To bridge this annotation gap, we introduce a lightweight translator, \emph{h36m2smpl}, comprising 6 GCN layers, 6 Transformer layers, and an MLP regression head. It achieves an MPJPE-24 of \textbf{32.2}\,mm and a PA-MPJPE-24 of \textbf{20.4}\,mm on the Human3.6M test set. While the absolute accuracy has room for improvement, the relative results suffice for downstream HMR comparison. Note that the standard HMR protocol measures MPJPE-24 via SMPL parameter regression followed by forward kinematics, whereas our 32.2\,mm is obtained from direct 17-to-24 joint regression, which means the two are not directly comparable.

\subsection{More Granularity Quantitative Results}
\label{subsec:more_granularity_results}
To further validate the generalizability of GPT, we retrain the decoders solely for the 21- and 25-joint poses, achieving 2D Mean Errors of \textbf{3.73}\,px and \textbf{4.66}\,px on the Human3.6M GT test set, respectively. These results are consistent with both the GT quality and the ratio of unseen joints, compared to other novel granularities.

\begin{table}[t]
  \centering
  \caption{Comparison with AugLift.}
  \label{tab:comparison_with_auglift}
  \setlength{\tabcolsep}{5pt}
  \resizebox{\columnwidth}{!}{
  \begin{tabular}{lccc}
    \toprule
    Method & Plug-in Type & Dataset & MPJPE $\downarrow$ \\
    \midrule
    PoseFormer($f=9$) & Baseline & H36M(CPN) & 50.5 \\
    AugLift & Depth Injection & H36M(CPN) & 49.8($\Delta0.7$) \\
    \midrule
    MixSTE($f=81$) & Baseline & H36M(GT) & 25.9 \\
    VARPose & Dense Information & H36M(GT) & 25.0($\Delta0.9$) \\
    \bottomrule
  \end{tabular}
  }
\end{table}

\subsection{Input-Enrichment Comparison}
As shown in Table~\ref{tab:comparison_with_auglift}, AugLift~\cite{warner2025auglift} ($\Delta0.7$) and VARPose ($\Delta0.9$) achieve comparable performance gains via fundamentally distinct input-enrichment paths. 
Specifically, AugLift injects image-level depth priors, whereas VARPose focuses on densifying the kinematic structure. 
Notably, AugLift generalizes better on the OOD 3DPW dataset, which is likely attributed to its robust RGB-derived depth cues. 
These observations suggest that the two methods play complementary roles in 3D pose estimation, and integrating them could potentially yield further improvements.

\section{More Ablation Studies}
\label{sec:more_ablation}
\paragraph{Ablation on GPT Modules} 
To analyze the effectiveness of each module involved in the quantization process, including area-based interpolation, residual quantization strategy, and MLP ResNet $\phi$, we conduct ablation studies with results presented in Table~\ref{tab:ablation_core_and_hyper}. Crucially, area-based interpolation serves as an indispensable non-learnable adapter for dimension transformation across varying quantization densities $\rho_k$ and cannot be removed. Replacing it with nearest-neighbor yields a 0.05 px degradation (0.41 vs. 0.36). Similarly, removing the MLP ResNet $\phi$ increases error by 0.16 px (0.52 vs. 0.36), and omitting residual quantization causes a substantial 1.29 px degradation (1.65 vs. 0.36), collectively confirming that all proposed components contribute meaningfully to the overall performance.

\begin{table}[t]
  \centering
  \caption{Ablation study on core components and hyperparameters.}
  \label{tab:ablation_core_and_hyper}
  \setlength{\tabcolsep}{3pt}
  \resizebox{\columnwidth}{!}{
  \begin{tabular}{lc|lc}
    \toprule
    Configuration& 
    2D Mean Error (px) $\downarrow$& Configuration & 2D Mean Error (px) $\downarrow$ \\
    \midrule
    \multicolumn{2}{l|}{\textit{Codebook Reconstruction}} & \multicolumn{2}{l}{\textit{Hyperparameter Selection}} \\
    Interp. Nearest & 0.41($\Delta$0.05) & \multicolumn{2}{l}{Loss Weight ($\lambda$)} \\
    w/o Res. Quant.      & 1.65 ($\Delta$1.29)  & 100     & 1.38 ($\Delta$1.02) \\
    w/o MLP ResNet $\phi$       & 0.52 ($\Delta$0.16)  & \graycell{\textbf{200}} & \graycell{\textbf{0.36 ($\Delta$0.00)}} \\
    \graycell{\textbf{Ours}} & \graycell{\textbf{0.36 ($\Delta$0.00)}} & 400     & 0.77 ($\Delta$0.41) \\
    \midrule
    \multicolumn{2}{l|}{\textit{Densification}} & \multicolumn{2}{l}{Quant. Density} \\
    w/o RoPE             & 8.90 ($\Delta$1.09)  & $\{102, 288, 576\}$ & 1.20 ($\Delta$0.84) \\
    \graycell{\textbf{Ours}} & \graycell{\textbf{7.81 ($\Delta$0.00)}} & \graycell{\textbf{Ours}} & \graycell{\textbf{0.36 ($\Delta$0.00)}} \\
    \bottomrule
  \end{tabular}
  }
\end{table}
\begin{figure}[t]
\centering
\includegraphics[width=\linewidth]{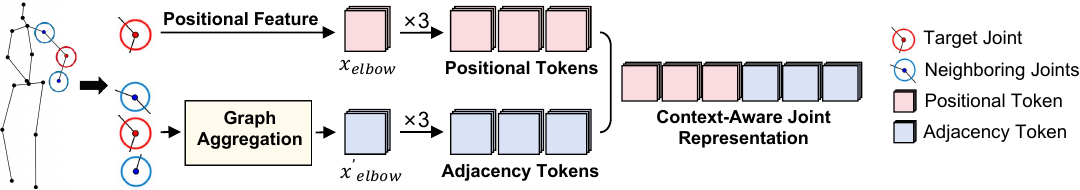}
\caption{Joint-to-Token Expansion via Feature Aggregation.}
\Description{Joint-to-Token Expansion via Feature Aggregation.}
\label{fig:joint-to-token_expansion}
\end{figure}

\paragraph{Selection of $\lambda$ and Density Scheduling}
The value $\lambda=200$ is chosen to balance the magnitudes of $\mathcal{L}_{\text{Recon}}$ ($\approx 6\times10^{-7}$) and $\mathcal{L}_{\text{VQ}}$ ($\approx 2\times10^{-4}$), yielding strong performance among tested values. For density scheduling, anchor densities $\{102, 288, 576\}$ are derived from training joint counts while intermediate densities $\{48, 192, 432\}$ serve as exact midpoints to ensure smooth transitions. As shown in Table~\ref{tab:ablation_core_and_hyper}, $\lambda=200$ achieves strong reconstruction quality, and removing intermediate densities degrades performance by 0.84~px, collectively validating both hyperparameter choices.

\paragraph{Joint-to-Token Expansion Strategy} 
\label{Joint-to-Token_Expansion_Strategy} To maintain the structural integrity of the skeleton, we utilize a Joint-to-Token expansion strategy, as shown in Figure~\ref{fig:joint-to-token_expansion}. Specifically, for a joint represented by 6 tokens, the module separates them into 3 positional tokens capturing self-properties and 3 adjacency tokens capturing context from the skeletal graph. During decoding, the features from these two groups are averaged independently and then summed to produce the final output feature. This ensures that the final regressed coordinates $\hat{\mathbf{x}}^\dagger$ are constrained by both individual joint accuracy and global structural coherence.

We conduct ablation studies to validate our Joint-to-Token expansion strategy by comparing two settings: (1) Position-Only: replicating only self-position information; (2) Balanced (ours): incorporating both self-position and neighborhood information. As shown in Table~\ref{tab:ablation_position_only} and Table~\ref{tab:ablation_neighborhood}, the Position-Only strategy degrades performance from 2.70~px (1-to-1) to 4.29~px (1-to-6), confirming that redundant tokens without neighborhood information harm generalization. In contrast, our Balanced strategy consistently improves performance as token count increases, achieving 0.36~px with the 1-to-6 configuration. These results validate that neighborhood information is critical for skeletal structure understanding. Moreover, within the balanced framework, increased token redundancy enriches the feature representation to better capture complex spatial dependencies.

\paragraph{Impact of Codebook Size and Embedding Dimension}
We analyze the sensitivity of codebook vocabulary size $V$ and embedding dimension $D$ on the Human3.6M dataset, as summarized in Table~\ref{tab:ablation_codebook_size_and_embedding}. We observe that increasing the vocabulary size $V$ from 512 to 4096 leads to a significant reduction in reconstruction error. Notably, as shown in the top half of Table~\ref{tab:ablation_codebook_size_and_embedding}, MACs and Params remain invariant to changes in $V$. This is because we employ Exponential Moving Average (EMA) to update the codebook, and the embedding vectors are registered as non-trainable \textit{buffers} rather than optimization parameters. In contrast, the embedding dimension $D$ serves as the primary driver of model complexity. As $D$ increases from 16 to 512, the MACs escalate significantly from $0.24$ G to $27.99$ G, as it directly dictates the width of the hierarchical encoders and decoders. While $D=512$ achieves the lowest error ($0.21$ px), it incurs a prohibitive computational cost. Therefore, we select $V=4096$ and $D=128$ as our final setting to achieve an optimal balance between reconstruction precision and hardware efficiency.

\begin{table}[t]
\centering
\caption{Effect of increasing the number of tokens for position-only token expansion.}
\label{tab:ablation_position_only}
\resizebox{\columnwidth}{!}{
\begin{tabular}{cccc}
\toprule
Tokens per Joint & MACs (G) & Params (M) &  2D Mean Error (px) \\
\midrule
1-to-1 & 0.27 & 3.84 & 2.70 \\
1-to-3 & 1.10 & 6.87 & 2.91 \\
1-to-6 & 3.06 & 17.09 & 4.29 \\
\bottomrule
\end{tabular}
}
\end{table}

\begin{table}[t]
\centering
\caption{Ablation study on token expansion strategies incorporating neighborhood information.}
\label{tab:ablation_neighborhood}
\resizebox{\columnwidth}{!}{
\begin{tabular}{cccc}
\toprule
Tokens per Joint & MACs (G) & Params (M) &  2D Mean Error (px) \\
\midrule
1-to-2 & 0.63 & 4.98 & 1.77 \\
1-to-4 & 1.66 & 9.52 & 0.91 \\
\graycell{\textbf{1-to-6 (Ours)}} & \graycell{3.06} & \graycell{17.09} & \graycell{\textbf{0.36}} \\
\bottomrule
\end{tabular}
}
\end{table}

\paragraph{Necessity of Using RoPE}
We quantitatively evaluate the impact of Rotary Position Embeddings (RoPE) on our framework's performance. As demonstrated in Table~\ref{tab:ablation_core_and_hyper}, the full VARPose model achieves a 2D Mean Error of 7.81~px. When RoPE is removed, the error significantly increases to 8.90~px, representing a substantial performance degradation of 1.09~px. This performance gap underscores that explicitly modeling relative positional relationships is crucial for structured skeletal data, as it allows the model to better capture the intricate topological dependencies between joints across different scales.

\begin{table}[t]
\centering
\caption{Model complexity under varying vocabulary size and embedding dimension.}
\label{tab:ablation_codebook_size_and_embedding}
\small
\renewcommand{\arraystretch}{1.2}
\begin{tabular}{lccccc}
\toprule
Setting & $V$ & $D$ & MACs (G) & Params (M)  & 2D Mean Error (px) \\
\midrule
\multirow{5}{*}{Vocab}
& 512  & 128 & 3.06 & 16.56 & 1.14 \\
& 1024 & 128 & 3.06 & 16.56 & 0.74 \\
& 2048 & 128 & 3.06 & 16.56 & 0.47 \\
& \graycell{4096} & \graycell{128} & \graycell{3.06} & \graycell{16.56} & \graycell{0.36} \\
& 8192 & 128 & 3.06 & 16.56 & 0.43 \\
\midrule
\multirow{4}{*}{Dim}
& 4096 & 16  & 0.24 & 13.68 & 2.55 \\
& 4096 & 64  & 1.20 & 14.37 & 1.20 \\
& \graycell{4096} & \graycell{128} & \graycell{3.06} & \graycell{16.56} & \graycell{0.36} \\
& 4096 & 512 & 27.99 & 60.17 & 0.21 \\
\bottomrule
\end{tabular}
\end{table}

\begin{figure}[t]
\centering
\includegraphics[width=\linewidth]{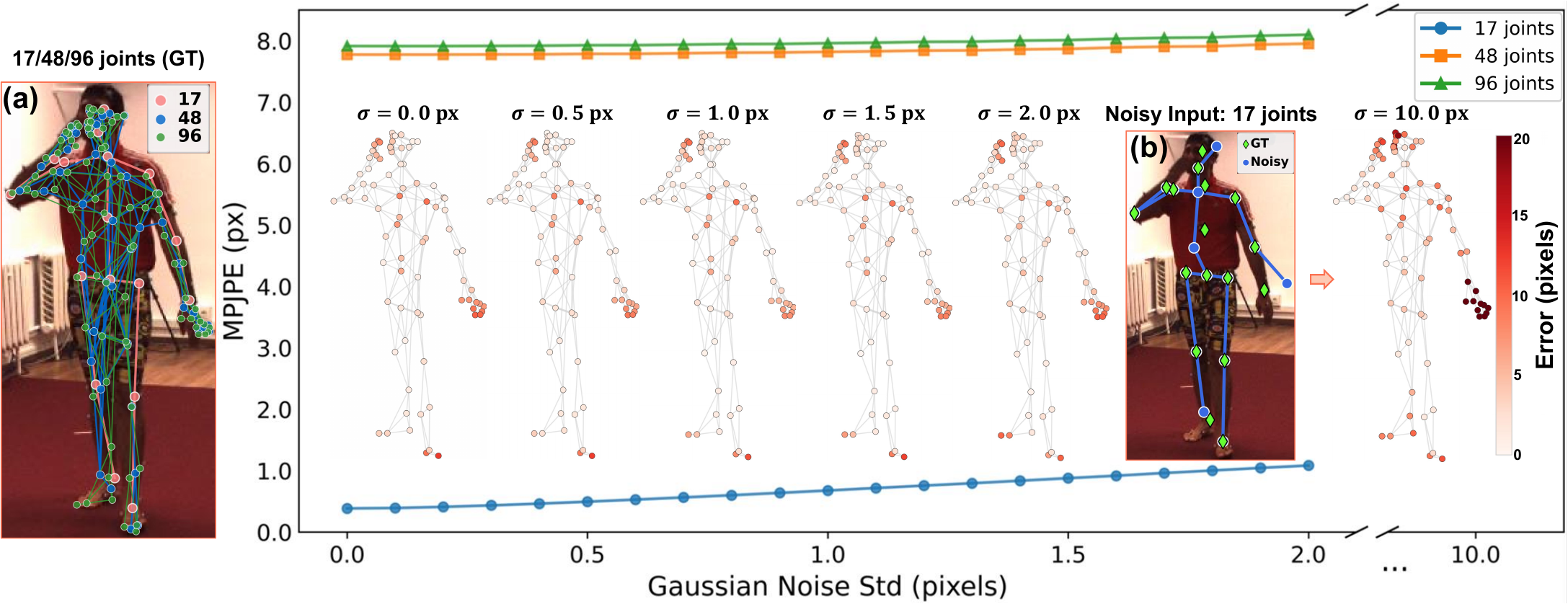}
\caption{Robustness analysis under various Gaussian noise.} 
\Description{Robustness analysis under various Gaussian noise.}
\label{fig:input_noise}
\end{figure}

\begin{figure}[t]
  \centering
  \begin{minipage}[t]{0.47\columnwidth}
    \centering
    \includegraphics[width=\linewidth]{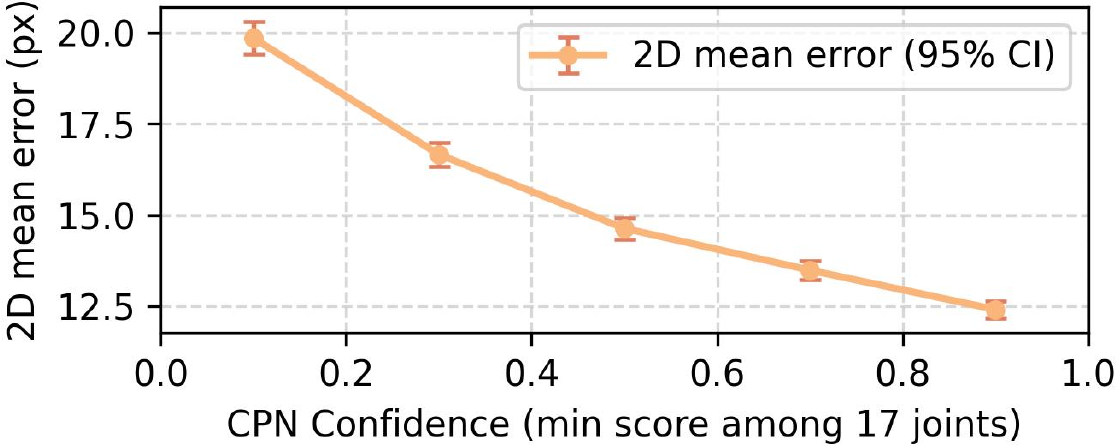}
    \captionsetup{font=footnotesize}
    \caption{CPN confidence vs. 2D mean error.}
    \Description{CPN confidence vs. 2D mean error.}
    \label{fig:conf_vs_error}
  \end{minipage}
  \hfill
  \begin{minipage}[t]{0.47\columnwidth}
    \centering
    \includegraphics[width=\linewidth]{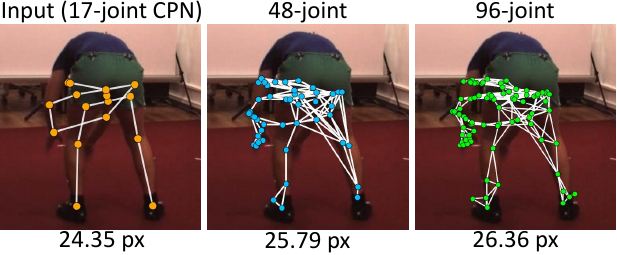}
    \captionsetup{font=footnotesize}
    \caption{Extreme-pose performance.}
    \Description{Extreme-pose performance.}
    \label{fig:extreme_pose}
  \end{minipage}
\end{figure}

\begin{figure}[t]
    \centering
    \includegraphics[width=\linewidth]{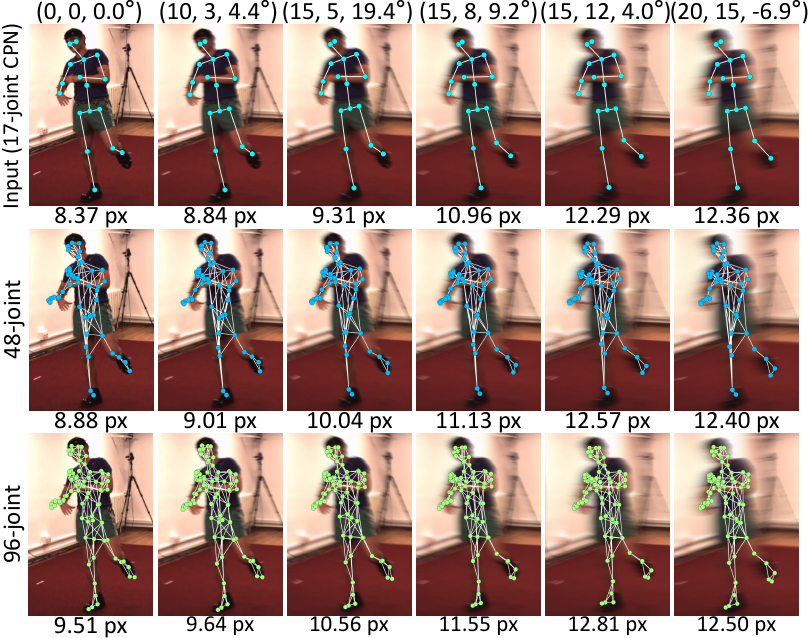}
    \captionsetup{font=footnotesize}
    \caption{Robustness under motion blur.}
    \Description{Robustness under motion blur.}
    \label{fig:motion_blur}
\end{figure}

\paragraph{More Robustness Experiments}
\label{subsec:more_robustness_experiments}
\emph{(a) Gaussian noise.} We inject Gaussian noise $N(0,\sigma^2)$ into 30\% of the GT 17-joint input pose $x_1$ before UniSkelar. As shown in Figure~\ref{fig:input_noise}, VARPose maintains structurally coherent outputs even under severe perturbations ($\sigma\!\!=\!\!10$\,px). 
\emph{(b) Real-world degradations.} We use \hl{non-dataset-adapted CPN} detections to evaluate three stress conditions: \textit{(i) \underline{Detection errors}:} Figure~\ref{fig:conf_vs_error} shows that the error increases controllably from 12.40 to 19.86\,px without model collapse as confidence decreases. \textit{(ii) \underline{Extreme poses}:} Figure~\ref{fig:extreme_pose} shows that VARPose preserves anatomical topology under severe bending and corrupted inputs. \textit{(iii) \underline{Motion blur} \cite{hendrycks2019robustness}:} Figure~\ref{fig:motion_blur} reports a maximum error of 12.81\,px across six severity levels. 
\emph{(c) Dataset-adapted detections.} Under the standard \hl{dataset-adapted CPN} protocol for 3D HPE, Section~\ref{subsec:more_2d_densification_results} reports a 2D error of \textbf{11.16}\,px, only $\Delta$3.35\,px above the GT-input setting. Together, these results demonstrate VARPose's robustness to synthetic perturbations and real-world input degradations.

\section{More Visualization Analysis}
\label{sec:more_visualization}
\subsection{Qualitative Results of 2D Hierarchical Reconstruction}
\begin{figure*}[t]
    \centering
    \includegraphics[width=0.85\linewidth]{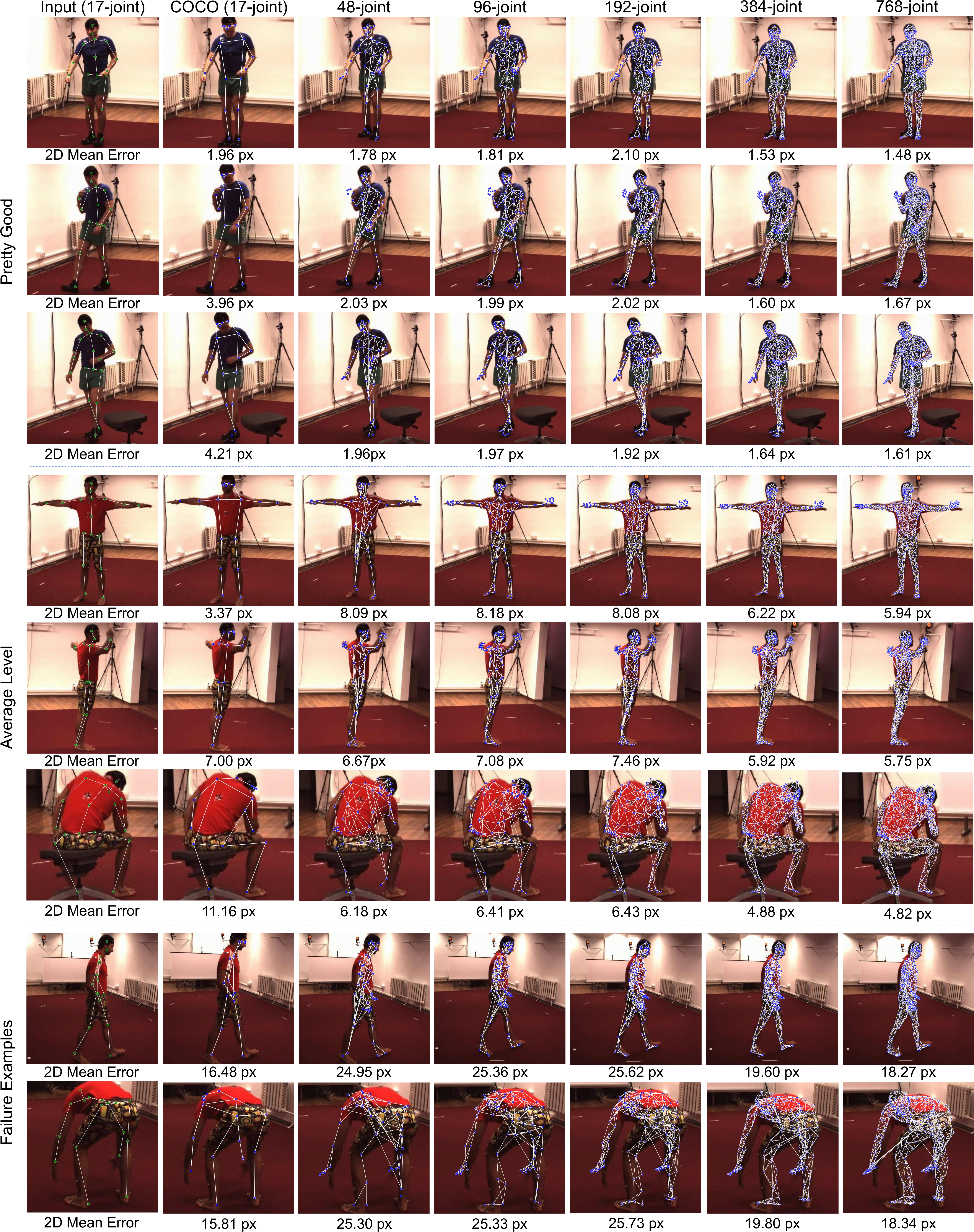}
    \caption{Qualitative analysis of VARPose across multiple granularities on the Human3.6M dataset. Each row displays the same frame across varying joint granularities, ranging from sparse (17-joint) to extremely dense (768-joint). The samples are organized vertically into ``Pretty Good'' (top), ``Average Level'' (middle), and ``Failure Examples'' (bottom) categories based on global error ranking. The corresponding 2D Mean Error (pixels) is provided below each sub-image. }
    \Description{Qualitative analysis of VARPose across multiple granularities on the Human3.6M dataset. Each row displays the same frame across varying joint granularities, ranging from sparse (17-joint) to extremely dense (768-joint). The samples are organized vertically into ``Pretty Good'' (top), ``Average Level'' (middle), and ``Failure Examples'' (bottom) categories based on global error ranking. The corresponding 2D Mean Error (pixels) is provided below each sub-image. }
    \label{fig:2d_demo_h36m_multiscale}
\end{figure*}

To qualitatively evaluate the performance of VARPose across different levels of detail, we visualize the 2D reconstruction results for six granularities ranging from COCO 17-joint skeletons to dense 768-joint skeletons. As illustrated in Figure~\ref{fig:2d_demo_h36m_multiscale}, despite the dramatic increase in the number of joints, the generated dense skeletons (192 to 768 joints) accurately follow the underlying body structure defined by the sparse 17-joint input. This suggests that the model effectively learns a continuous representation of the human body manifold, allowing it to ``super-resolve'' a sparse skeleton into a dense silhouette without losing structural integrity. In complex poses involving significant self-occlusion or bending, the model's error naturally increases. However, even in these challenging cases, VARPose exhibits graceful degradation. While the specific joint locations might shift, the generated dense point clouds still maintain a coherent human-like shape rather than collapsing into random noise. For instance, in the extreme bending case (bottom row), the 768-joint output still successfully reconstructs the overall volume and orientation of the torso and limbs, which is highly beneficial for downstream tasks like mesh recovery or action recognition.

\subsection{Qualitative Results of 3D Downstream Tasks}
\begin{figure*}[t]
    \centering
    \includegraphics[width=0.8\linewidth]{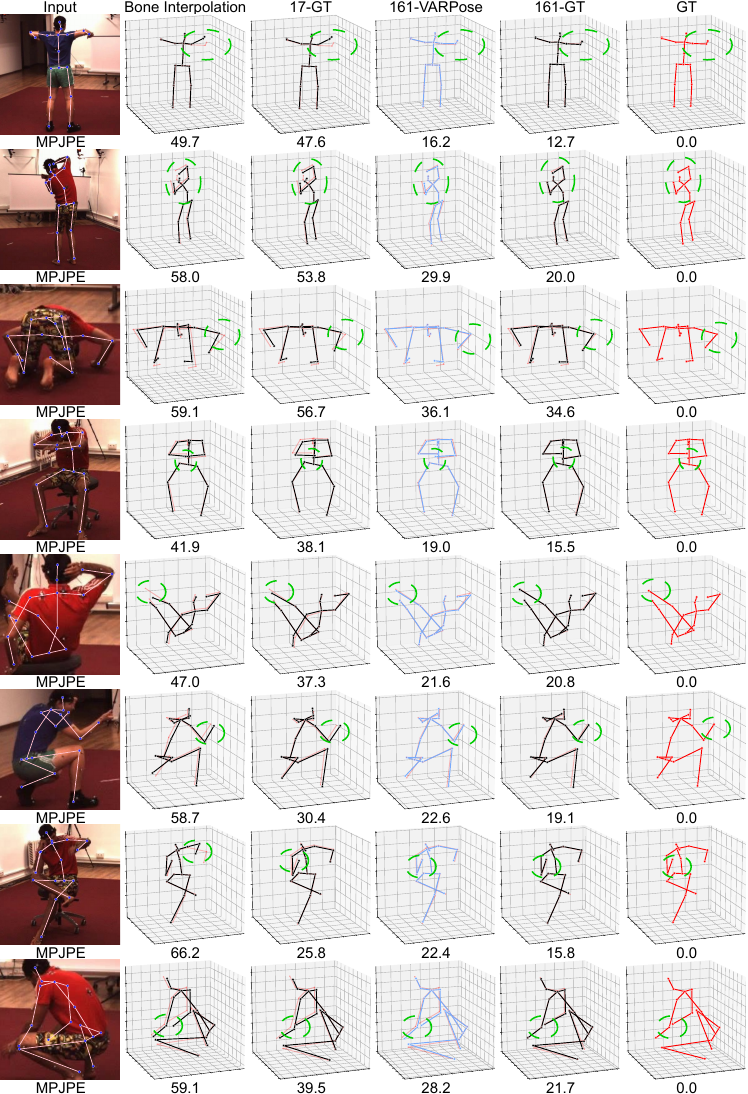}
    \caption{Qualitative results of 3D HPE. We use MixSTE as the lifting backbone with the same experimental settings described in the ablation study. The predicted 2D keypoints are directly lifted to 3D joint coordinates via MixSTE.}
    \Description{Qualitative results of 3D HPE. We use MixSTE as the lifting backbone with the same experimental settings described in the ablation study. The predicted 2D keypoints are directly lifted to 3D joint coordinates via MixSTE.}
    \label{fig:visualization_of_hpe}
\end{figure*}

\begin{figure*}[t]
    \centering
    \includegraphics[width=0.9\linewidth]{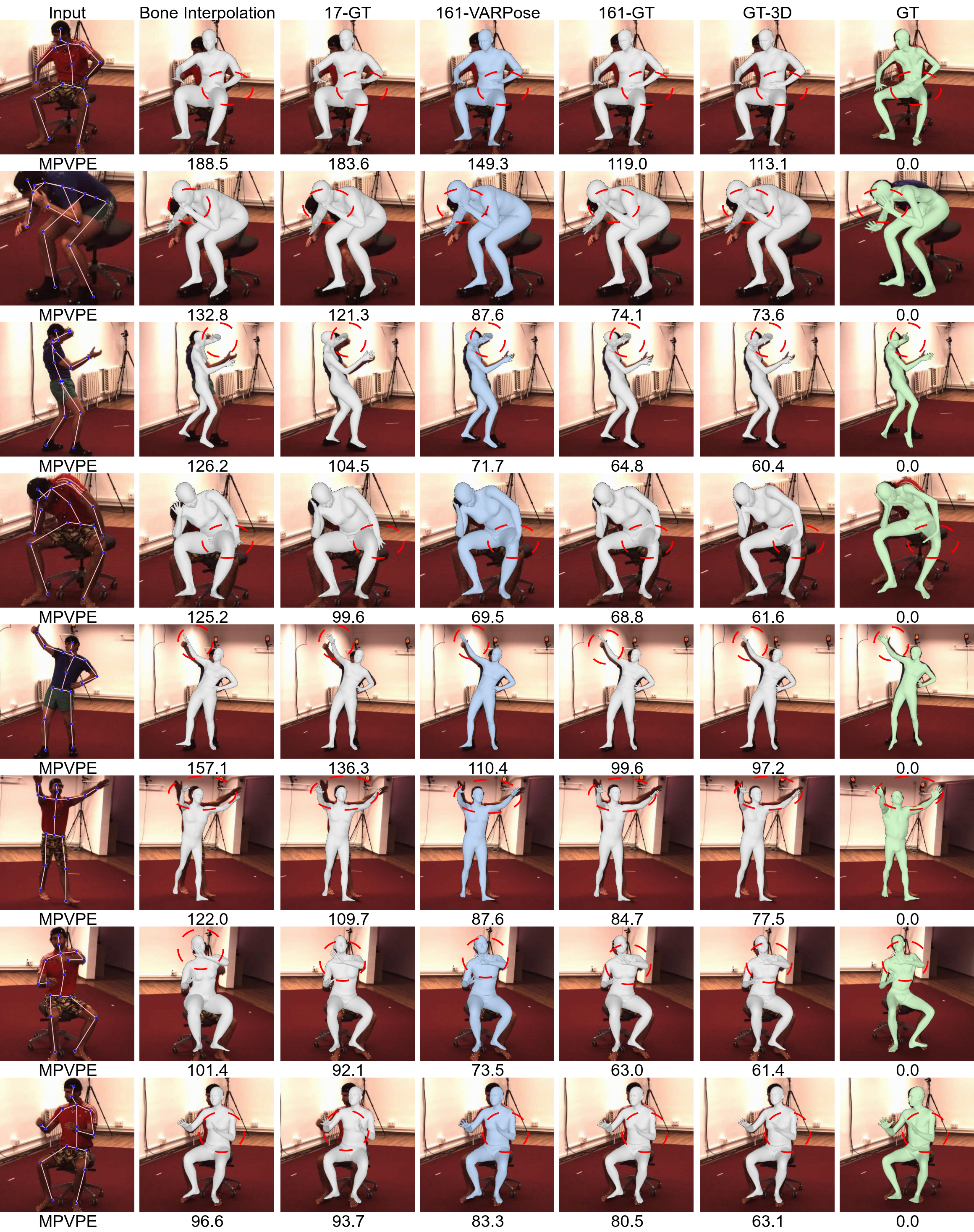}
    \caption{Qualitative results of 3D HMR. We use MixSTE as the lifting backbone with the same experimental settings described in the ablation study. The predicted 2D keypoints are first lifted to 3D H36M joint coordinates via MixSTE, then regressed to SMPL mesh through H36M-to-SMPL parameter fitting.}
    \Description{Qualitative results of 3D HMR. We use MixSTE as the lifting backbone with the same experimental settings described in the ablation study. The predicted 2D keypoints are first lifted to 3D H36M joint coordinates via MixSTE, then regressed to SMPL mesh through H36M-to-SMPL parameter fitting.}
    \label{fig:visualization_of_hmr}
\end{figure*}

As shown in Figure~\ref{fig:visualization_of_hpe}, we adopt MixSTE~\cite{zhang2022mixste} as the lifting model, consistent with the ablation setting in the main text. When fed dense inputs derived from mesh vertices, MixSTE leverages richer anatomical information, leading to enhanced robustness against self-occlusion and superior accuracy in limb regions. This validates our motivation for densification. The downstream HMR task follows the same trend, as illustrated in Figure~\ref{fig:visualization_of_hmr}, further underscoring the effectiveness of VARPose's densification strategy.

\section{Limitations and Future Work}
\label{sec:limitation}
The current approach has three main limitations.
\textit{First}, our densification does not explicitly model temporal consistency since frame-to-frame jitter from mesh annotations and independent per-frame processing may limit effectiveness in downstream 3D tasks requiring smooth motion priors, which motivates future integration of temporal modules into the representation-projection paradigm.
\textit{Second}, the current plug-in strategy treats dense keypoints as augmented inputs rather than learning unified spatial structures, yielding suboptimal efficiency-performance trade-offs that a future end-to-end framework leveraging the codebook's granularity-agnostic nature could resolve.
\textit{Third}, performance degradation on non-rigid body parts is inherent to the SMPL-based representation, while degradation in multi-person scenarios stems from the single-person modeling paradigm. Together, these define the fundamental scope boundaries of this work.

\end{document}